%% file: root.tex
\documentclass{article}
\usepackage{amssymb}
\input{format/packages}
\input{format/macros}

\usepackage[final]{corl_2025} 

\title{\textsc{MEReQ}: Max-Ent Residual-Q Inverse RL for Sample-Efficient Alignment from Intervention}

\author{
    \textbf{Yuxin Chen}$^{\ast,1}$, \textbf{Chen Tang}$^{\ast,2}$, \textbf{Jianglan Wei}$^{1}$, \textbf{Chenran Li}$^1$, 
    \textbf{Ran Tian}$^1$,\\
    \textbf{Xiang Zhang}$^1$, \textbf{Wei Zhan}$^{1}$, \textbf{Peter Stone}$^{2,3}$, \textbf{Masayoshi Tomizuka}$^{1}$\\
    $^{\ast}$Denotes equal contribution \\
    $^1$\textit{University of California, Berkeley} \quad $^2$\textit{The University of Texas at Austin} \quad $^3$\textit{Sony AI}
}

\input{notation_macros}

\begin{document}
\maketitle


\begin{abstract}
Aligning robot behavior with human preferences is crucial for deploying embodied AI agents in human-centered environments. A promising solution is interactive imitation learning from human intervention, where a human expert observes the policy's execution and provides interventions as feedback. However, existing methods often fail to utilize the prior policy efficiently to facilitate learning, thus hindering sample efficiency. In this work, we introduce Maximum-Entropy Residual-Q Inverse Reinforcement Learning (\textsc{MEReQ})\footnote{
Website:
\url{https://thomaschen98.github.io/mereq.github.io/}.
}, designed for sample-efficient alignment from human intervention. Instead of inferring the complete human behavior characteristics, \textsc{MEReQ} infers a \emph{residual reward function} that captures the discrepancy between the human expert's and the prior policy's underlying reward functions. Residual Q-Learning (RQL) is then employed to align the policy with human preferences using the inferred reward function. Extensive evaluations on simulated and real-world tasks show that \textsc{MEReQ} achieves sample-efficient alignment from human intervention compared to baselines.
\end{abstract}


\section{Introduction}
	
\label{sec:introduction}
Recent progress in embodied AI has enabled advanced robots to perform complex real-world tasks that go beyond pre-scripted routines and highly controlled environments. Increasing research attention has been focused on how to align their behavior with human preferences~\citep{ji2023ai, arzate2020survey}, which is crucial for their deployment in human-centered environments. One promising approach is interactive imitation learning, where a pre-trained policy can interact with a human and align its behavior to the human's preference through human feedback~\citep{arzate2020survey,cui2021understanding}. In this work, we focus on interactive imitation learning using \emph{human interventions} as feedback. In this setting, the human expert observes the policy during task execution and intervenes whenever it deviates from their preferred behavior. A straightforward approach~\cite{kelly2019hg,liu2023robot,zhang2016query} is to update the policy through behavior cloning (BC)~\citep{ross2010efficient}\textemdash maximizing the likelihood of the collected intervention samples under the learned policy distribution. However, BC ignores the sequential nature of decision-making, leading to compounded errors~\citep{garg2021iq}. Additionally,~\citet{jiang2024transic} pointed out that these approaches are not ideal for the fine-tuning setting, 
since the policies are fine-tuned to fit solely the collected intervention data, thus suffering from catastrophic forgetting, which hinders sample efficiency. 

\begin{figure*}[t]
     \centering
    \includegraphics[width=0.8\linewidth]{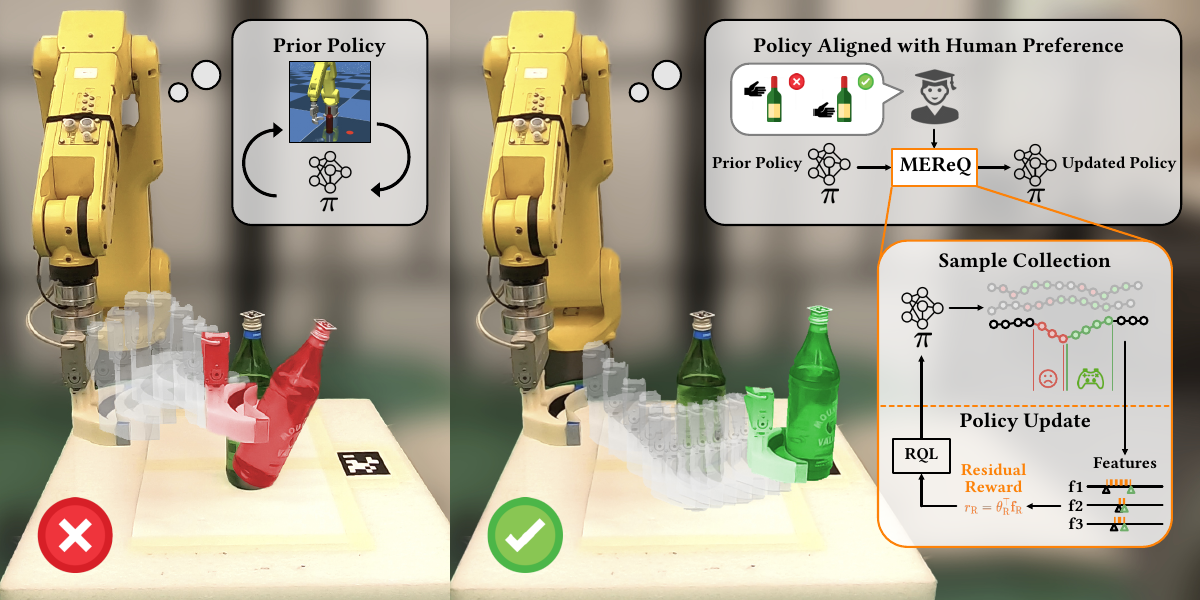}
    \caption{{\bf Overview of \textsc{MEReQ}}, designed for sample-efficient alignment from human intervention. From human intervention samples, \textsc{MEReQ} infers a residual reward that captures the discrepancy between the human expert’s and the prior policy’s underlying reward functions via maximum-entropy inverse reinforcement and then updates the prior policy with Residual Q-Learning (RQL).}
    \label{fig:motivation}
\end{figure*}

We instead study the learning-from-intervention problem within the inverse reinforcement learning (IRL) framework~\cite{ng2000algorithms,ziebart2008maximum}. IRL models the expert as a sequential decision-making agent who maximizes cumulative returns based on their internal reward function, and infers this reward function from expert demonstrations. IRL inherently accounts for the sequential nature of human decision-making and the effects of transition dynamics~\citep{arora2021survey}. Maximum-entropy IRL (MaxEnt-IRL) further accounts for the sub-optimality in human behavior by favoring randomness in the policy that is learned from the inferred reward function~\citep{von1947theory,baker2007goal,ziebart2008maximum}. However, directly applying IRL to fine-tune a prior policy from interventions can still be inefficient. The prior policy is ignored in the learning process, except as an initialization for the learning policy. Consequently, like other approaches, it fails to effectively leverage the prior policy to reduce the number of intervention samples needed.

To address this shortcoming, this paper introduces \textbf{M}aximum-\textbf{E}ntropy \textbf{Re}sidual-\textbf{Q} Inverse Reinforcement Learning (\textsc{MEReQ}) for \emph{sample-efficient alignment from human intervention}. The key insights behind \textsc{MEReQ} are to infer a \emph{residual reward function} that captures the discrepancy between the human expert's internal reward function and that of the prior policy, rather than inferring the full human reward function from interventions. \textsc{MEReQ} then employs Residual Q-Learning (RQL)~\citep{li2024residual} to fine-tune and align the policy with the unknown expert reward, leveraging the inferred residual reward function. We show that \textsc{MEReQ} can effectively align a prior policy with fewer interventions than baselines in both simulation and real-world tasks.

\section{Related Work}
\label{sec:related work}
Interactive imitation learning (IL) utilizes human feedback to align policies with human behavior preference~\cite {arzate2020survey,cui2021understanding}. Forms of human feedback include preference~\citep{yue2012k,jain2013learning, christiano2017deep, biyik2022learning, lee2021pebble, wang2022skill,ouyang2022training,myers2023active,rafailov2024direct,hejna2023contrastive,tian2023matters}, interventions~\citep{zhang2016query, saunders2017trial, wang2021appli, celemin2019interactive, peng2024learning, kelly2019hg, mandlekar2020human, spencer2020learning}, scaled feedback~\citep{knox2012reinforcement,argall2010tactile,fitzgerald2019human,bajcsy2017learning,najar2020interactively,wilde2021learning,warnell2018deep,macglashan2017interactive} and rankings~\citep{brown2019extrapolating}. Like ours, several approaches~\citep{tian2023matters,cui2018active,spencer2022expert,bobu2018learning,bobu2021feature} opt to infer the internal reward function of humans from their feedback and update the policy using the inferred reward function. While these methods have demonstrated improved performance and sample efficiency compared to those without a human in the loop~\cite {liu2023robot}, further enhancing efficiency beyond the sample collection pattern has not been thoroughly explored. Our method addresses this gap by building on a prior policy and inferring only the residual reward to improve sample efficiency. Based on similar motivation, Jiang et al. introduced TRANSIC~\cite {jiang2024transic}, which learns a residual policy from human corrections and integrates it with the prior policy for autonomous execution. Their framework addresses the sim-to-real gaps through human corrections, whereas our method targets the problem where there is a mismatch in rewards between the prior policy and the human subject. 

\section{Preliminaries}
\label{sec:preliminaries}
In this section, we briefly review two techniques used in \textsc{MEReQ}, which are RQL and MaxEnt-IRL, to establish the foundations for the main technical results. 

\subsection{Policy Customization and Residual Q-Learning}
\label{sec:policy_customization}
Li et al.~\citep{li2024residual} introduced a new problem setting termed \emph{policy customization}. Given a prior policy, the goal is to find a new policy that jointly optimizes 1) the task objective the prior policy is designed for; and 2) additional task objectives specified by a downstream task. The authors proposed RQL as an initial solution. Formally, RQL assumes the prior policy $\pi:\mathcal{S}\times\mathcal{A}\mapsto[0,\infty)$ is a max-ent policy solving a Markov Decision Process (MDP) defined by the tuple $\mathcal{M}=(\mathcal{S},\mathcal{A},r,p,\rho_0,\gamma)$, where $\mathcal{S}\in\mathbb{R}^S$ is the state space, $\mathcal{A}\in\mathbb{R}^A$ is the action space, $r:\mathcal{S}\times\mathcal{A}\mapsto\mathbb{R}$ is the reward function, $p:\mathcal{S}\times\mathcal{A}\times\mathcal{S}\mapsto[0,\infty)$ represents the probability density of the next state $\mathbf{s}_{t+1}\in\mathcal{S}$ given the current state $\mathbf{s}_{t}\in\mathcal{S}$ and action $\mathbf{a}_{t}\in\mathcal{A}$, $\rho_0$ is the starting state distribution, and $\gamma\in[0,1)$ is the discount factor. That is to say, $\pi$ follows the Boltzmann distribution~\citep{haarnoja2017reinforcement}:
\begin{equation}
\label{eq:boltzmann_dist}
    \pi(\mathbf{a}|\mathbf{s})=\frac{1}{Z_\mathrm{s}}\exp\left(\frac{1}{\alpha}Q^\star(\mathbf{s},\mathbf{a})\right),
\end{equation}
where $Q^\star(\mathbf{s},\mathbf{a})$ is the soft $Q$-function as defined in~\citep{haarnoja2017reinforcement}, which satisfies the soft Bellman equation.

Policy customization is then formalized as finding a max-ent policy $\hat{\pi}: \mathcal{S}\times\mathcal{A}\mapsto[0,\infty)$ for a new MDP defined by $\hat{\mathcal{M}} = (\mathcal{S}, \mathcal{A}, r + r_\mathrm{R}, p, \rho_0, \gamma)$, where $r_\mathrm{R}: \mathcal{S} \times \mathcal{A} \mapsto \mathbb{R}$ is a \emph{residual reward} function that quantifies the discrepancy between the original task objective and the customized task objective for which the policy is being customized. Given $\pi$, RQL finds this customized policy without knowledge of the prior reward $r$. Specifically, define the soft Bellman update operator~\cite{haarnoja2017reinforcement,haarnoja2018soft} as:
\begin{equation}
\begin{aligned}
\label{eq:soft_bellman_updata}
\hat{Q}_{t+1}(\mathbf{s},\mathbf{a}) = r_\mathrm{R}(\mathbf{s},\mathbf{a})+r(\mathbf{s},\mathbf{a})
+\gamma\mathbb{E}_{\mathbf{s}'\sim p(\cdot|\mathbf{s},\mathbf{a})}\left[\hat\alpha\log\int_\mathcal{A}\exp\left(\frac{1}{\hat{\alpha}}\hat{Q}_t(\mathbf{s}',\mathbf{a}')\right)d\mathbf{a}'\right],
\end{aligned}
\end{equation}
where $\hat{Q}_t$ is the estimated soft $Q$-function at the $t^\mathrm{th}$ iteration. RQL introduces a \emph{residual} $Q$-function defined as $Q_{\mathrm{R},t}:=\hat{Q}_t-Q^\star$. It was shown that $Q_{\mathrm{R},t}$ can be learned without knowing $r$ and the customized policy can be defined using ${Q}_{\mathrm{R},t}$ and $\pi$. 


RQL considers the case where $r_\mathrm{R}$ is specified. In this work, we aim to customize the policy towards a human behavior preference, under the assumption that  $r_\mathrm{R}$ is unknown a priori. \textsc{MEReQ} is proposed to infer $r_\mathrm{R}$ from interventions and customize the policy towards the inferred residual reward. 

\subsection{Maximum-Entropy Inverse Reinforcement Learning}
\label{sec:maximum_entropy_inverse_reinforcement_learning}
In the IRL setting, an agent is assumed to optimize a reward function defined as a linear combination of a set of \emph{features} $\mathbf{f}:\mathcal{S}\times\mathcal{A}\mapsto\mathbb{R}^f$ with weights $\theta\in\mathbb{R}^f$: $r = \theta^\top\mathbf{f}(\zeta)$. Here $\mathbf{f}(\zeta)$ is the trajectory \emph{feature counts}, $\mathbf{f}(\zeta) = \sum_{(\mathbf{s}_i,\mathbf{a}_i)}\mathbf{f}(\mathbf{s}_i,\mathbf{a}_i)$, which are the sum of the state-action features $\mathbf{f}(\mathbf{s}_i,\mathbf{a}_i)$ along the trajectory $\zeta$. IRL~\cite {ng2000algorithms} aligns the feature expectations between an observed expert and the learned policy. However, multiple reward functions can yield the same optimal policy, and different policies can result in identical feature counts~\cite {ziebart2008maximum}. One way to resolve this ambiguity is by employing the principle of maximum entropy~\citep{jaynes1957information}, where policies that yield equivalent expected rewards are equally probable, and those with higher rewards are exponentially favored:
\begin{equation}
\begin{aligned}
\label{eq:maximum_entropy_IRL_policy}
p(\zeta|\theta)=\frac{p(\zeta)}{Z_\zeta(\theta)}\exp\left(\theta^\top\mathbf{f}(\zeta)\right)=\frac{p(\zeta)}{Z_\zeta(\theta)}\exp\left[\sum_{(\mathbf{s}_i,\mathbf{a}_i)}\theta^\top\mathbf{f}(\mathbf{s}_i,\mathbf{a}_i)\right],
\end{aligned}
\end{equation}
where $Z_\zeta(\theta)$ is the \emph{partition function} defined as $\int p(\zeta)\exp\left(\theta^\top\mathbf{f}(\zeta)\right) d\zeta$ and $p(\zeta)$ is the trajectory prior. The optimal weight $\theta^\star$ is obtained by maximizing the likelihood of the observed data:
\begin{equation}
\label{eq:maximum_entropy_IRL_weight}
\theta^\star = \arg\max_\theta\mathcal{L} = \arg\max_\theta\log p(\tilde{\zeta}|\theta),
\end{equation}
where $\tilde{\zeta}$ represents the demonstration trajectories. The optima can be obtained using gradient-based optimization with gradient defined as $\nabla_\theta\mathcal{L} = \mathbf{f}(\tilde{\zeta})-\int p(\zeta|\theta)\mathbf{f}(\zeta)d\zeta$. At the maxima, the feature expectations align, ensuring that the learned policy's performance matches the demonstration, regardless of the specific reward weights the agent aims to optimize.

\section{Problem Formulation}
\label{sec:problem_formulation}
\vspace{-5pt}
We focus on the problem of aligning a given prior policy with human behavior preferences by learning from \emph{human intervention}. In this setting, a human expert observes a policy as it executes the task and intervenes whenever the policy behavior deviates from the expert's preference. The expert then continues executing the task until they are comfortable disengaging. Formally, we assume access to a prior policy $\pi$ to execute, which is an optimal max-ent policy with respect to an unknown reward function $r$. We assume a human with an internal reward function $r_\mathrm{expert}$ that differs from $r$ observes $\pi$'s execution and provides interventions. The problem objective is to infer $r_\mathrm{expert}$ and use the inferred reward function to learn a policy $\hat{\pi}$ that matches the max-ent optimal policy with respect to $r_\mathrm{expert}$. During learning, we can execute the updated policy under human supervision to collect new intervention samples. However, we want to minimize the number of samples collected, so as to limit the cognitive cost to the human. We assume access to a simulator. 

If the ground truth $r_\mathrm{expert}$ were known, we could synthesize the max-ent optimal policy with respect to that reward using max-ent RL~\cite{haarnoja2017reinforcement,haarnoja2018soft}. We could then evaluate the success of a particular method by measuring how closely the learned policy $\hat{\pi}$ approximates this optimal policy. However, we cannot access the human's internal reward function in practice. Therefore, we assess the effectiveness of an approach by the human intervention rate during policy execution, measured as the fraction of time steps during which the human intervenes in a task episode. We aim to develop an algorithm to learn a policy that can minimize the number of intervention samples required to reach a target intervention rate threshold. Additionally, we design synthetic tests where we know the expert reward and train a max-ent policy under the ground-truth reward as a human proxy, so that we can directly measure the sub-optimality of the learned policy (see Sec.~\ref{sec:experiments}).
\section{Max-Ent Residual-Q Inverse Reinforcement Learning (\textsc{MEReQ})}
\label{sec:mereq}
\vspace{-5pt}
In this section, we present \textsc{MEReQ}, a sample-efficient algorithm for alignment from human intervention. We first introduce a naive MaxEnt-IRL solution (Sec.~\ref{sec:naive-solution}), and analyze its drawbacks to motivate residual reward learning (Sec.~\ref{sec:residual_reward_inference}).  We then present the complete algorithm (Sec.~\ref{sec:mereq_algorithm}). 

\subsection{A Naive Maximum-Entropy IRL Solution}\label{sec:naive-solution}

A naive way to solve the target problem is to directly apply MaxEnt-IRL to infer the human reward function $r_\mathrm{expert}$ and find $\hat\pi$. We model the human expert with the widely recognized model of Boltzmann rationality~\citep{von1947theory,baker2007goal}, which conceptualizes human intent through a reward function and portrays humans as choosing trajectories proportionally to their exponentiated rewards~\citep{bobu2020less}. We model $r_\mathrm{expert}$ as a linear combination of features, as stated in Sec.~\ref{sec:maximum_entropy_inverse_reinforcement_learning}. We initialize the learning policy $\hat{\pi}$ as the prior policy $\pi$. We then iteratively collect human intervention samples by executing $\hat{\pi}$, and then infer $r_\mathrm{expert}$ and update $\hat{\pi}$ based on the collected intervention samples. We refer to this solution as \textbf{MaxEnt-FT}, with FT denoting fine-tuning. In our experiments, we also study a variation with randomly initialized $\hat{\pi}$, denoted as \textbf{MaxEnt}.

In each sample collection iteration $i$, {MaxEnt-FT} executes the current policy $\hat{\pi}$ for $T$ timesteps under human supervision. The single rollout of length $T$ is split into two classes of segments depending on who takes control, which are policy segments $\xi^\mathrm{p}_1$, $\xi^\mathrm{p}_2$, $\dots$, $\xi^\mathrm{p}_m$, and expert segments $\xi^\mathrm{e}_1$,  $\xi^\mathrm{e}_2$, $\dots$, $\xi^\mathrm{e}_n$, where a segment $\xi$ is a sequence of state-action pairs $\xi=\left\{(\mathbf{s}_1,\mathbf{a}_1),\dots,(\mathbf{s}_j,\mathbf{a}_j)\right\}$. We define the collected \emph{policy trajectory} in this iteration as the union of all policy segments, $\Xi^\mathrm{p}=\bigcup_{k=1}^m\xi^\mathrm{p}_k$. Similarly, we define the \emph{expert trajectory} as $\Xi^\mathrm{e}=\bigcup_{k=1}^n\xi_k^\mathrm{e}$. Note that $\sum_{k=1}^m |\xi^\mathrm{p}_k|+\sum_{k=1}^n |\xi_k^\mathrm{e}|=T$.

Under the Boltzmann rationality model, each expert segment follows the distribution in Eqn.~\eqref{eq:maximum_entropy_IRL_policy}. Assuming the expert segments are all independent from each other, the likelihood of the expert trajectory can be written as $p(\Xi^\mathrm{e}|\theta)=\prod_{k=1}^n p(\xi^\mathrm{e}_k|\theta)$. We can then infer the weights of the unknown human reward function by maximizing the likelihood of the observed expert trajectory, that is
\begin{equation}\label{eqn:reward-inference}
\theta^\star = \arg\max_\theta\log p(\Xi^\mathrm{e}|\theta) = \arg\max_\theta\sum_{k=1}^n\log p(\xi_k^\mathrm{e}|\theta),
\end{equation}
then update $\hat{\pi}$ to be the max-ent optimal policy with respect to the reward function ${\theta^\star}^\top \mathbf{f}$. Note that directly optimizing these reward inference and policy update objectives completely disregards the prior policy. Thus, this naive solution is inefficient in the sense that it is expected to require many human interventions, as it overlooks the valuable information embedded in the prior policy. 

\subsection{Residual Reward Inference and Policy Update}
\label{sec:residual_reward_inference}
In this work, we aim to develop an alternative algorithm that can utilize the prior policy to solve the target problem in a sample-efficient manner. We start with reframing the policy update step in the naive solution as a \emph{policy customization} problem~\citep{li2024residual}. Specifically, we can rewrite the unknown human reward function as the sum of $\pi$'s underlying reward function $r$ and a \emph{residual reward} function $r_\mathrm{R}$. We expect some feature weights to be zero for $r_\mathrm{R}$, specifically for the reward features for which the expert’s preferences match those of the prior policy. Thus, we represent $r_\mathrm{R}$ as a linear combination of the non-zero weighted feature set $\mathbf{f}_\mathrm{R}:\mathcal{S}\times\mathcal{A}\mapsto\mathbb{R}^{f_\mathrm{R}}$ with weights $\theta_\mathrm{R}$. 

If $\theta_\mathrm{R}$ is known, we can apply RQL to update the learning policy $\hat\pi$ without knowing $r$ (see Sec.~\ref{sec:policy_customization}). Yet, $\theta_\mathrm{R}$ is unknown, and MaxEnt can only infer the full reward weights $\theta$ (see Eqn.~\eqref{eqn:reward-inference}). Instead, we introduce a novel method that enables us to \emph{directly infer the residual weights $\theta_\mathrm{R}$ from expert trajectories without knowing $r$}, and then apply RQL with $\pi$ and $r_R$ to update the policy $\hat\pi$, which will be more sample-efficient than the naive solution, MaxEnt. The residual reward inference method is derived as follows. By substituting the residual reward function into the maximum likelihood objective function, we obtain the following objective function:
\begin{equation}
\mathcal{L}=\sum_{k=1}^n\left[r(\xi^\mathrm{e}_k)+\theta_\mathrm{R}^\top\mathbf{f}_\mathrm{R}(\xi^\mathrm{e}_k)\right]-\log Z_{k}(\theta_\mathrm{R}),
\end{equation}
where $\mathbf{f}_\mathrm{R}(\xi)$ is a shorthand for $\sum_{(\mathbf{s}_i,\mathbf{a}_i)\in\xi}\mathbf{f}_\mathrm{R}(\mathbf{s}_i,\mathbf{a}_i)$, and $r(\xi)$ is a shorthand for $\sum_{(\mathbf{s}_i,\mathbf{a}_i)\in\xi}r(\mathbf{s}_i,\mathbf{a}_i)$. The partition function $Z_{k}$ is defined as $Z_{k}(\theta_\mathrm{R})=\int p(\xi_k)\exp\left[r(\xi_k)+\theta_\mathrm{R}^\top\mathbf{f}_\mathrm{R}(\xi_k)\right]d\xi_k$, with $|\xi_k|=|\xi^\mathrm{e}_k|$ for each $k$. We can then derive the gradient of the objective function as:
\begin{equation}
\label{eq:theta_gradient_1}
\begin{aligned}
\nabla_{\theta_\mathrm{R}}\mathcal{L} &= \sum_{k=1}^n\mathbf{f}_\mathrm{R}(\xi^\mathrm{e}_k)- \sum_{k=1}^n\frac{1}{Z_k(\theta_\mathrm{R})}\int p(\xi_k)\exp\left[r(\xi_k)+\theta_\mathrm{R}^\top\mathbf{f}_\mathrm{R}(\xi_k)\right]\mathbf{f}_\mathrm{R}(\xi_k)d\xi_k,\\
&= \sum_{k=1}^n\mathbf{f}_\mathrm{R}(\xi^\mathrm{e}_k) - \sum_{k=1}^n\mathbb{E}_{\xi_k\sim p(\xi_k|\theta_\mathrm{R})}\left[\mathbf{f}_\mathrm{R}(\xi_k)\right].
\end{aligned}
\end{equation}
The second term is essentially the expectation of the feature counts of $\mathbf{f}_\mathrm{R}$ under the soft optimal policy under the current $\theta_\mathrm{R}$. Therefore, we approximate the second term with samples obtained by rolling out the current policy $\hat\pi$ in the simulation environment: 
\begin{equation}\label{eqn:gradient}
\sum_{k=1}^n\mathbb{E}_{\xi_k\sim p(\xi_k|\theta_\mathrm{R})}\left[\mathbf{f}_\mathrm{R}(\xi_k)\right]\approx \frac{1}{T}\sum_{k=1}^n|\xi_k^\mathrm{e}|\cdot \mathbb{E}_{\xi\sim\hat\pi(\xi)}\left[\mathbf{f}_\mathrm{R}(\xi)\right].
\end{equation}

The term $\nicefrac{|\xi_k^\mathrm{e}|}{T}$ is introduced to match the rollout length with that of the expert intervention samples. We can then apply gradient descent to infer $\theta_\mathrm{R}$ directly, without inferring the prior reward term $r$. 


\subsection{Algorithm}
\label{sec:mereq_algorithm}
The complete \textsc{MEReQ} algorithm is shown in Algorithm~\ref{alg:mereq}. In summary, \textsc{MEReQ} consists of an outer loop for sample collection and an inner loop for policy updates. In each sample collection iteration $i$, \textsc{MEReQ} runs the current policy $\hat\pi$ under the supervision of a human expert, collecting policy trajectory $\Xi^\mathrm{p}_i$ and expert trajectory $\Xi^\mathrm{e}_i$ (Line 3). Afterward, \textsc{MEReQ} enters the inner policy update loop to update the policy using the collected samples, \textit{i.e.}, $\Xi^\mathrm{p}_i$ and $\Xi^\mathrm{e}_i$, during which the policy is rolled out in a simulation environment to collect samples for reward gradient estimation and policy training. Concretely, each policy update iteration $j$ alternates between applying a gradient descent step with step-size $\eta$ to update the residual reward weights $\theta_\mathrm{R}$ (Line 10), where the gradient is estimated (Line 7) following Eqn.~\eqref{eq:theta_gradient_1} and Eqn.~\eqref{eqn:gradient}, and applying RQL to update the policy using $\pi$ and the updated $\theta_\mathrm{R}$ (Line 11). The inner loop is terminated when the residual reward gradient is smaller than a certain threshold $\epsilon$ (Line 8-9). The outer loop is terminated when the expert intervention rate, denoted by $\lambda$, hits a certain threshold $\delta$ (Line 4-5).

{\bf Pseudo Expert Trajectories.} Inspired by previous learning from intervention algorithms~\cite{mandlekar2020human,spencer2022expert}, we further categorize the policy trajectory $\Xi^\mathrm{p}_i$ into \emph{snippets} labeled as ``good-enough'' samples and ``bad'' samples. Let $\xi$ represent a single continuous segment within $\Xi^\mathrm{p}_i$, and let $[a,b)\circ\xi$ denote a \emph{snippet} of the segment $\xi$, where $a,b\in[0,1]$, $a\leq b$, referring to the snippet starting from the $\left[a|\xi|\right]$ timestep to the $\left[b|\xi|\right]$ timestep of the segment. The absence of intervention in the initial portion of $\xi$ implicitly indicates that the expert considers these actions satisfactory, leading us to classify the first $1-\kappa$ fraction of $\xi$ as ``good-enough'' samples. We aggregate all such ``good-enough'' samples to form what we term the \emph{pseudo-expert} trajectory, defined as $\Xi_i^+:=\{(\mathbf{s},\mathbf{a})|(\mathbf{s},\mathbf{a})\in[0,1-\kappa)\circ\xi,~\forall\xi\subset\Xi^\mathrm{p}_i\}$. Pseudo-expert samples offer insights into expert preferences without additional interventions. If \textsc{MEReQ} uses the pseudo-expert trajectory to learn the residual reward function, it is concatenated with the expert trajectory, resulting in an augmented expert trajectory set, $\Xi^\mathrm{e}_i=\Xi^\mathrm{e}_i\cup \Xi_i^+$, to replace the original expert trajectory. Adding these pseudo-expert samples only affects the gradient estimation step in Line 8 of Algorithm~\ref{alg:mereq}.

\vspace{-10pt} 
\begin{figure}[h]
\begin{algorithm}[H]
\caption{Learn Residual Reward Weights $\theta_\mathrm{R}$ in MEReQ-IRL Framework} 
\begin{algorithmic}[1]
    \Require $\pi$, $\delta$, $\epsilon$, $\mathbf{f}_\mathrm{R}$, and $\eta$
    \State $\theta_{\mathrm{R}}\leftarrow\mathbf{0}$, $\hat\pi\leftarrow \pi$
    \For{$i=0, \dots, N_\text{data}$}
        \State Execute current policy $\hat\pi$ under expert supervision to get $\Xi_{i}^\mathrm{e}$ and $\Xi^\mathrm{p}_{i}$
        \If {$\lambda_i=\texttt{len}(\Xi_i^\mathrm{e})/\texttt{len}(\Xi^\mathrm{p}_{i}+\Xi_{i}^\mathrm{e})<\delta$}
            \Comment{{\color{RoyalBlue}Intervention rate lower than threshold}}
            \State \Return
        \EndIf
        \For{$j=0, \dots, N_\text{update}$}
            \State Estimate the residual reward gradient $\nabla_{\theta_\mathrm{R}}\mathcal{L}$
            \If {$\nabla_{\theta_\mathrm{R}}\mathcal{L}<\epsilon$}
                \Comment{{\color{RoyalBlue}$\theta_{\mathrm{R}}$ converges}}
                \State \Return
            \EndIf
            \State $\theta_{\mathrm{R}}\leftarrow \theta_{\mathrm{R}}+\eta\nabla_{\theta_\mathrm{R}}\mathcal{L}$
            \State $\hat\pi\leftarrow\texttt{Residual\_Q\_Learning}(\pi, \hat\pi, \mathbf{f}_\mathrm{R}, \theta_{\mathrm{R}})$
        \EndFor
    \EndFor
\end{algorithmic}
\label{alg:mereq}
\end{algorithm}
\vspace{-15pt}
\end{figure}

\section{Experiments}
\label{sec:experiments}
\textbf{Tasks.} We design multiple simulated and real-world tasks, which are categorized into two settings depending on the expert type. First, we consider learning from a \emph{synthesized} expert. We specify a residual reward function and train an expert policy. Then, we define a \emph{heuristic-based} rule to decide when the expert should intervene or disengage. Since we know the expert policy, we can directly evaluate the sub-optimality of the learned policy. Under this setting, we consider four simulated tasks: 1) \textbf{\textit{Highway-Sim:}} The task is to control a vehicle to navigate through highway traffic in the \texttt{highway-env}~\citep{highway-env}. The prior policy can change lanes arbitrarily to maximize progress, while the expert policy encourages the vehicle to stay in the right-most lane; 2) \textbf{\textit{Bottle-Pushing-Sim:}} The task is to control a robot arm to push a wine bottle to a goal position in \texttt{MuJoCo}~\citep{todorov2012mujoco}. The prior policy can push the bottle anywhere along the height of the bottle, while the expert policy encourages pushing near the bottom of the bottle; 3) \textbf{\textit{Erasing-Sim:}} In this task, a robot arm erases a marker on a whiteboard in \texttt{MuJoCo}~\citep{todorov2012mujoco}. The prior policy applies insufficient force for effective erasing, whereas the expert encourages greater contact force to ensure the marker is fully erased; 4) \textbf{\textit{Pillow-Grasping-Sim:}} The task is to control a robot arm to grasp a pillow in \texttt{MuJoCo}~\citep{todorov2012mujoco}. The prior policy does not have a grasping point preference, whereas the expert favors grasping from the center.


We then validate \textsc{MEReQ} with \emph{human-in-the-loop} (HITL) experiments. The tasks are similar to the ones with synthesized experts, specifically: 1) \textbf{\textit{Highway-Human:}} A human expert monitoring task execution through a GUI and intervening using a keyboard. The human is instructed to keep the vehicle in the rightmost lane if possible; 2) \textbf{\textit{Bottle-Pushing-Human:}} This experiment is conducted on a Fanuc LR Mate 200$i$D/7L 6-DoF robot arm with a customized tooltip to push the bottle. The human expert intervenes with a SpaceMouse when the robot does not aim for the bottom of the bottle; 3) \textbf{\textit{Pillow-Grasping-Human:}} The experiment configuration is similar to bottle pushing, but the robot arm is equipped with a two-finger gripper. In these experiments, the specific algorithm variant was hidden from the expert during each trial. Please refer to Appendix~\ref{app:environment_settings} for more details.

\textbf{Baselines and Evaluation Protocol.}
We compare \textcolor{mereq}{\textbf{MEReQ}} with the following baselines: \textcolor{mereqnp}{\textbf{MEReQ-NP}}, a MEReQ variation that does not use pseudo-expert trajectories (\textit{i.e.}, \textbf{N}o \textbf{P}seudo); 2) \textcolor{maxentft}{\textbf{MaxEnt-FT}}, the naive max-ent IRL solution (see Sec.~\ref{sec:naive-solution}); 3) \textcolor{maxent}{\textbf{MaxEnt}}, the naive solution but with random policy initialization; 4) \textcolor{hgdagger}{\textbf{HG-DAgger-FT}}, a variant of DAgger tailored for interactive imitation learning (IL) from human experts in real-world systems~\cite{kelly2019hg}; 5) \textcolor{iwr}{\textbf{IWR-FT}}, an intervention-based behavior cloning method with intervention weighted regression~\cite{mandlekar2020human}. To ensure a fair comparison between \textcolor{mereq}{\textbf{MEReQ}} and the two interactive IL methods, we implemented the following adaptations: 1) We rolled out the prior policy to collect samples, which were then used to warm start \textcolor{hgdagger}{\textbf{HG-DAgger-FT}} and \textcolor{iwr}{\textbf{IWR-FT}} with behavior cloning. As shown in Fig.~\ref{fig:sample_efficiency} (top), the initial intervention rates of the warm-started \textcolor{hgdagger}{\textbf{HG-DAgger-FT}} and \textcolor{iwr}{\textbf{IWR-FT}} are comparable to those of the prior policy of \textcolor{mereq}{\textbf{MEReQ}}; 2) Since both interactive IL methods maintain a dataset of all collected expert samples, we retained the full set of expert trajectories from each iteration, $\Xi^\mathrm{e}=\bigcup_i\Xi^\mathrm{e}_i$, where $i$ denotes the iteration number, for the residual reward gradient calculation (Algorithm~\ref{alg:mereq}, line 7) of \textcolor{mereq}{\textbf{MEReQ}}. As discussed in Sec.~\ref{sec:problem_formulation}, we use expert intervention rate as the main criterion to assess policy performance. We are primarily interested in the \emph{sample efficiency} of the tested approaches. Specifically, we measure the number of expert samples required to have the intervention rate $\lambda$ reach a certain threshold value $\delta$. 


\subsection{Experimental Results}
\label{sec:simulation_results}

\begin{figure*}[t]
    \centering
    \includegraphics[width=\linewidth]{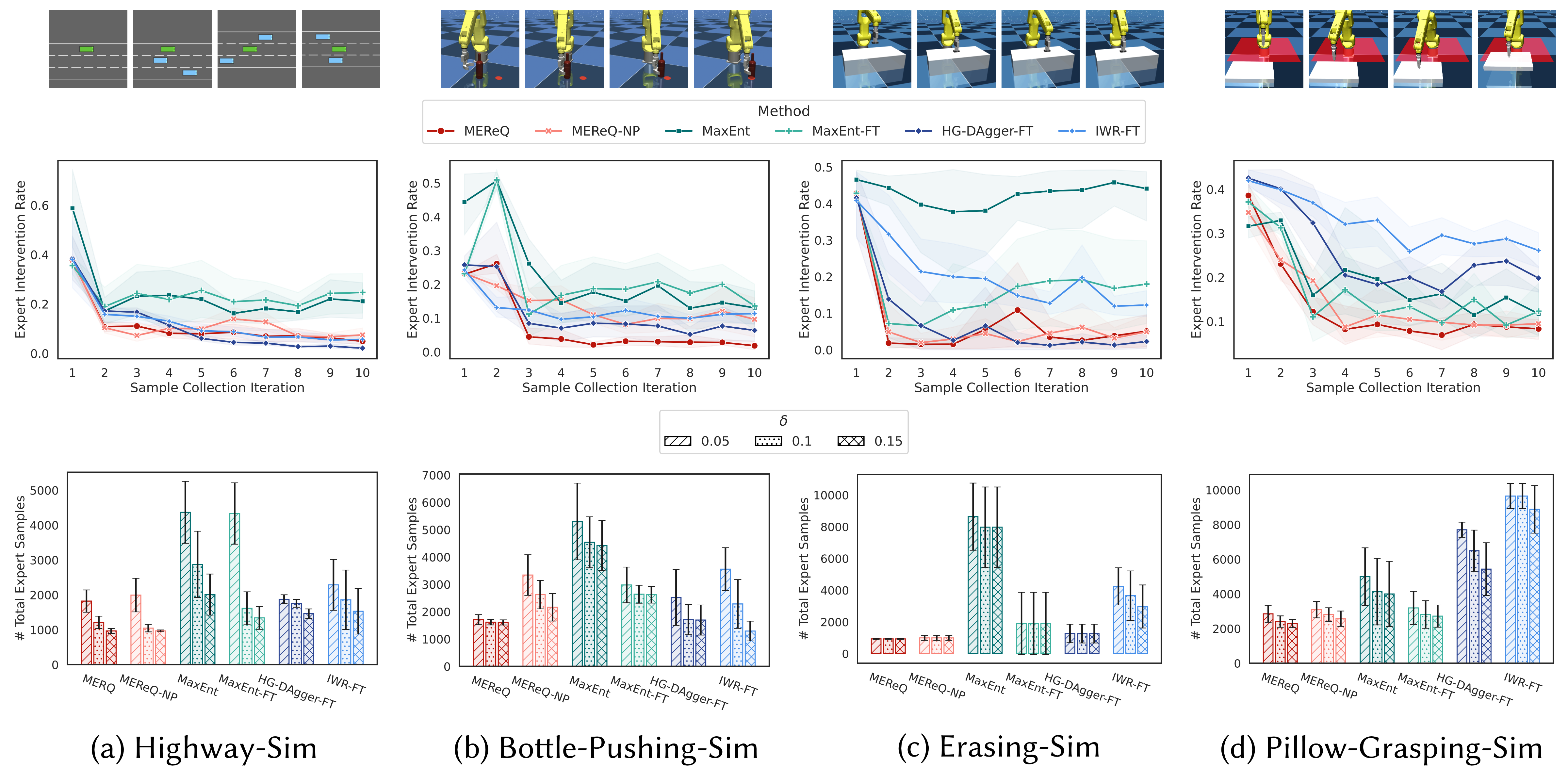}
    
    \caption{\textbf{Sample Efficiency.} \textbf{(Top)} \textcolor{mereq}{\textbf{MEReQ}} converges faster and maintains at low intervention rate throughout the sample collection iterations. The error bands indicate a 95\% confidence interval across 8 trials. \textbf{(Bottom)} \textcolor{mereq}{\textbf{MEReQ}} requires fewer total expert samples to achieve comparable policy performance compared to baselines under varying intervention rate thresholds $\delta$. The error bars indicate a 95\% confidence interval. See Tab.~\ref{tab:sample_efficiency} in Appendix~\ref{app:additional_results} for detailed values.}
    \label{fig:sample_efficiency}
\end{figure*}

{\bf Experiments with Synthesized Experts.} We evaluate each method using 8 random seeds and 10 data collection iterations per run. For each method, we compute the number of expert intervention samples needed to reach intervention rate thresholds $\delta = [0.05, 0.1, 0.15]$. As shown in Fig.~\ref{fig:sample_efficiency} (top), \textcolor{mereq}{\textbf{MEReQ}} consistently achieves higher sample efficiency than baselines across all tasks and thresholds. Notably, it exhibits significantly lower variance across seeds compared to \textcolor{hgdagger}{\textbf{HG-DAgger-FT}} and \textcolor{iwr}{\textbf{IWR-FT}}, especially in more challenging environments like \textit{Bottle-Pushing-Sim}, \textit{Erasing-Sim}, and \textit{Pillow-Grasping-Sim}. We further analyze behavior alignment in \textit{Bottle-Pushing-Sim}, and find that the \textcolor{mereq}{\textbf{MEReQ}} policy matches the synthesized expert more closely in terms of feature and reward distributions than baselines (see Appendix~\ref{app:additional_results} for more detailed results). 

We would like to highlight two design choices that enable \textcolor{mereq}{\textbf{MEReQ}}’s sample efficiency and stability. First, the combination of residual reward learning and RQL allows MEReQ to effectively leverage the prior policy to facilitate sample efficiency, thus outperforming \textcolor{maxentft}{\textbf{MaxEnt-FT}}. In particular, \textcolor{maxentft}{\textbf{MaxEnt-FT}}’s expert intervention rate quickly rises to match that of \textcolor{maxent}{\textbf{MaxEnt}} after the first iteration in \textit{Bottle-Pushing-Sim}, suggesting it only benefits from the prior policy in the early stage. Second, incorporating pseudo-expert samples further stabilizes training. \textcolor{mereq}{\textbf{MEReQ}} shows lower variance than \textcolor{mereqnp}{\textbf{MEReQ-NP}}. We hypothesize this because expert sample variance increases when intervention rates are low. Pseudo-expert samples help reduce the variance and stabilize training.

\begin{figure*}[t]
    \centering
    \includegraphics[width=\linewidth]{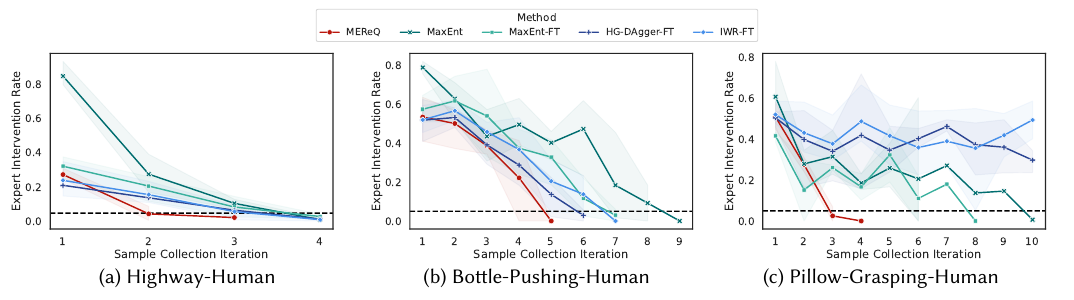}
    
    \caption{\textbf{Human Effort.} \textcolor{mereq}{\textbf{MEReQ}} can effectively reduce human efforts. The error bands indicate a 95\% confidence interval across 3 trials. See Tab.~\ref{tab:human_efforts} in Appendix~\ref{app:additional_results} for detailed values.}
    \label{fig:human_efforts}
\end{figure*}

\begin{figure}[t]
    \centering
    \begin{subfigure}[b]{0.5\linewidth}
        \centering
        \includegraphics[width=0.95\linewidth]{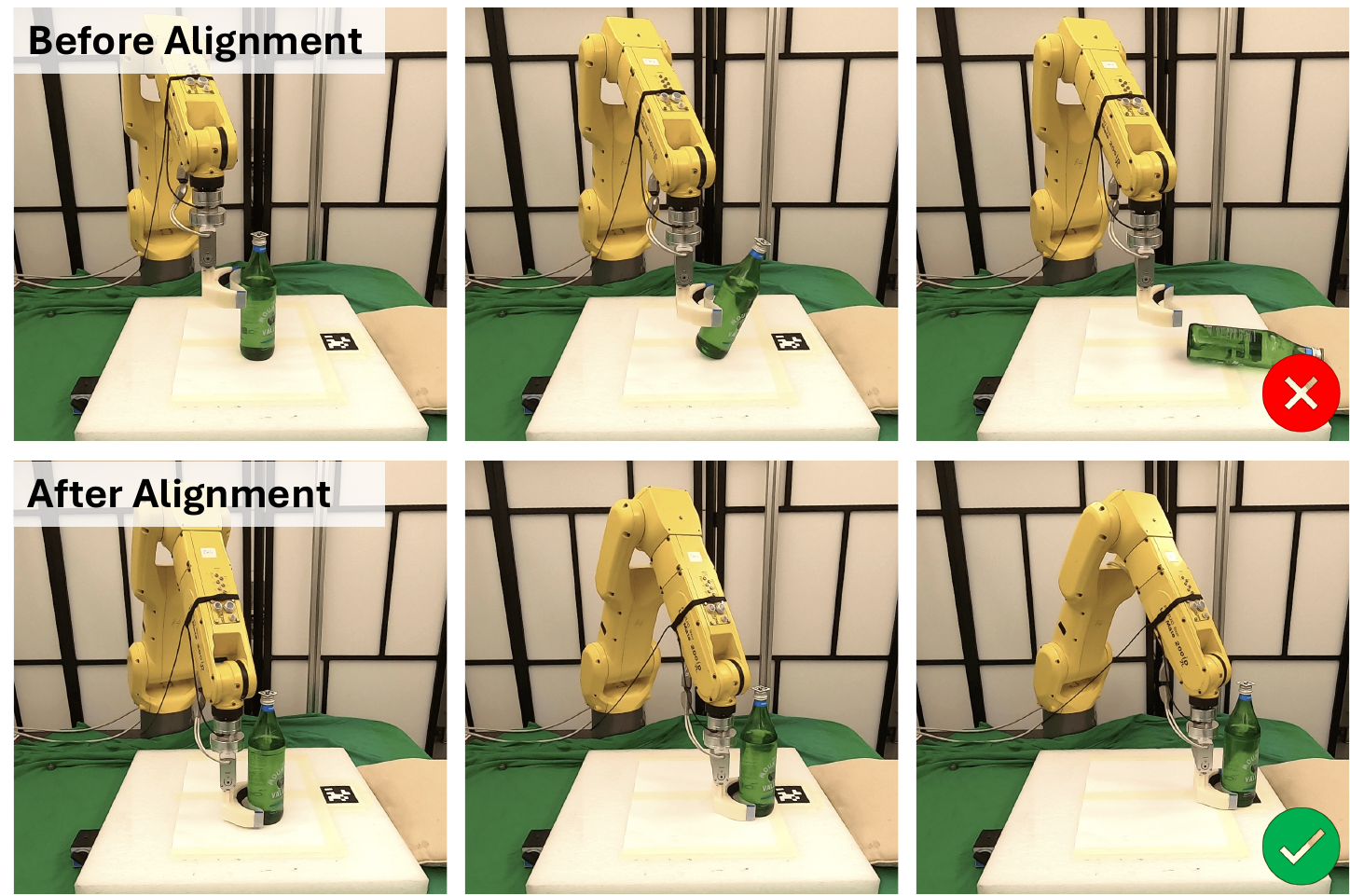}
    \end{subfigure}\hfill
    \begin{subfigure}[b]{0.5\linewidth}
        \centering
        \includegraphics[width=0.95\linewidth]{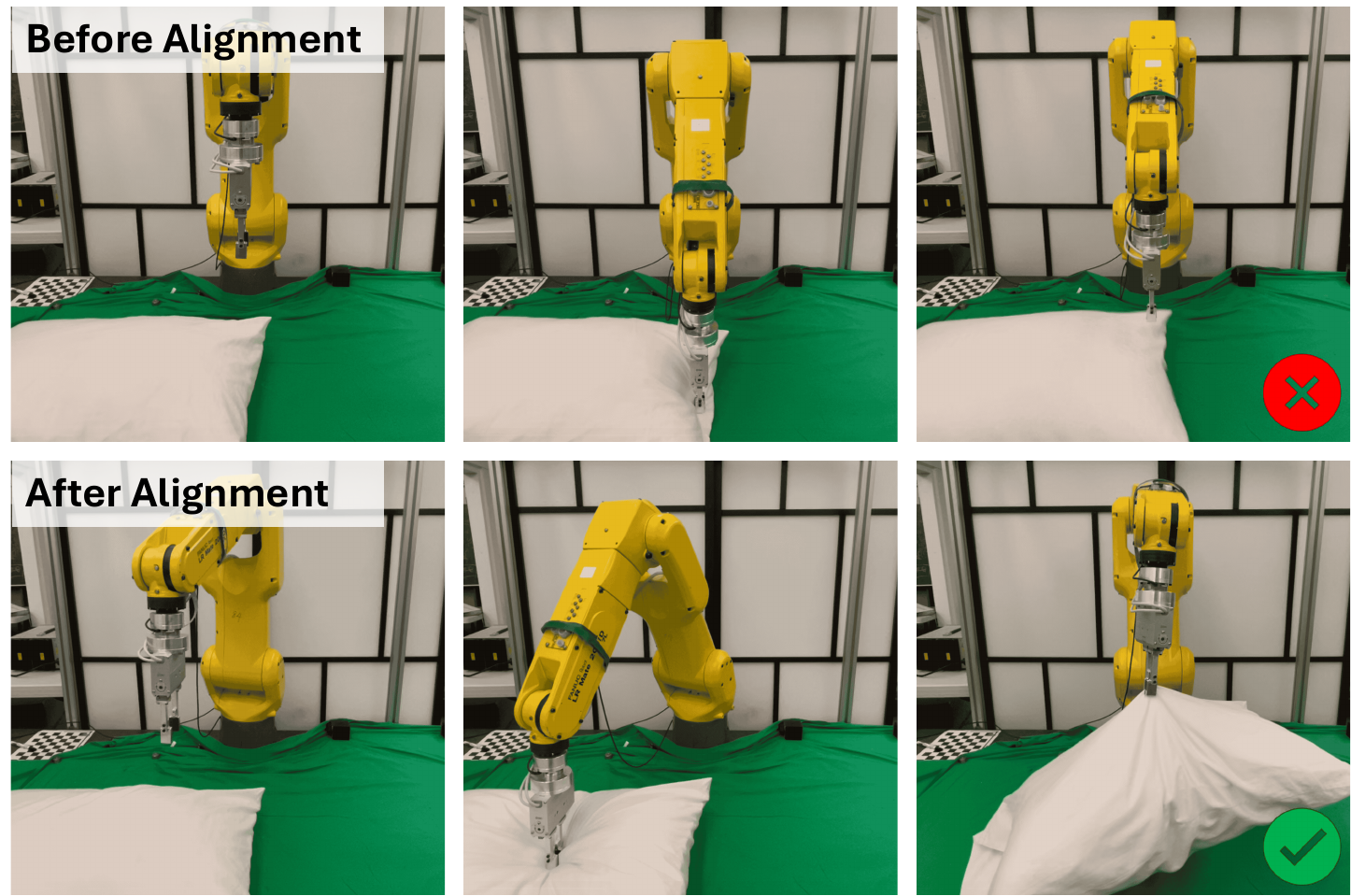}
    \end{subfigure}
    \caption{\textbf{Left: \textit{Bottle-Pushing-Human} Rollout.} Before alignment, the robot knocks down the bottle with a high contact point. The robot pushes the bottle to the goal position with low contact point after alignment. \textbf{Right: \textit{Pillow-Grasping-Human} Rollout.} Before alignment, the robot fails to grasp the pillow by the center. The robot grasps the pillow successfully after alignment.}
    \label{fig:robot_demo}
\end{figure}

{\bf Human-in-the-loop Experiments.} We investigate whether \textcolor{mereq}{\textbf{MEReQ}} effectively reduces the effort required from human experts. We set $\delta=0.05$ and conducted three trials per method with a human expert. In \textit{Highway-Human}, the human expert is tasked with supervising 50 rollouts, each consisting of 40 steps, during every outer loop. For both \textit{Bottle-Pushing-Human} and \textit{Pillow-Grasping-Human}, 10 rollouts are supervised by the expert in each outer loop. The training process concludes once the specified threshold is reached. As shown in Fig.~\ref{fig:human_efforts}, compared to the max-ent IRL baselines, \textcolor{mereq}{\textbf{MEReQ}} aligns the prior policy with human preferences in fewer sample collection iterations and with fewer human intervention samples (See Tab.~\ref{tab:human_efforts} in Appendix~\ref{app:additional_results}). These results demonstrate that \textcolor{mereq}{\textbf{MEReQ}} can be effectively adopted in real-world applications. As shown in Fig.~\ref{fig:robot_demo}, the prior policies fail to complete the \textit{Bottle-Pushing} or \textit{Pillow-Grasping} tasks due to inappropriate contact points, while the aligned policies successfully complete the tasks after adaptation from human interventions.



\section{Limitations and Future Work}
\label{sec:limitation}
We introduce MEReQ, a novel algorithm for sample-efficient policy alignment from human intervention, which learns a residual reward function capturing the discrepancy between the human expert's and the prior policy's rewards. Across seven tasks in both simulation and real-world systems, MEReQ achieves alignment with significantly fewer human interventions than baseline approaches. While these results highlight the effectiveness of MEReQ, several limitations and promising future directions remain.

First, the current policy-updating process requires rollouts in a simulation environment, causing delays between sample-collection iterations. Parallel rollouts could speed up the training process. Adopting offline or model-based RL could also be a promising direction. Second, high variance in expert intervention samples could perturb the stability of \textsc{MEReQ}'s training procedure. While the pseudo-expert approach can mitigate this issue, it is nevertheless a heuristic. More principled methods to reduce sample variance may be useful to  further improve \textsc{MEReQ}. Additionally, noise and inconsistency in intervention actions may also perturb performance. We report preliminary studies on these effects across different algorithms in App.~\ref{app:additional_results}, though how to fully address them is beyond the current scope and remains an important avenue for future research. 

Third, \textsc{MEReQ} follows the linear reward model commonly used in inverse reinforcement learning (IRL). We are actively exploring IRL methods without this assumption~\cite{levine2011nonlinear,boularias2011relative} and plan to extend MEReQ along this line in future work. Finally, in our human-in-the-loop experiments, each task was overseen by a single operator, which may introduce bias based on that person's skills, system familiarity, and tolerance level to undesirable behaviors. A broader study involving more participants would deepen our insight into how trust and subjectivity influence the timing, criteria, and frequency of interventions.



\bibliography{example}  
\input{appendix}

\end{document}

%% file: format/packages.tex
\usepackage{algorithm}
\usepackage{algorithmicx}
\usepackage{eqparbox}
\usepackage{subcaption}
\usepackage{caption}
\usepackage{multirow}
\usepackage[dvipsnames,usenames]{xcolor}
\usepackage[noend]{algpseudocode}
\usepackage{amsmath,amsfonts,amssymb}
\usepackage{booktabs}
\usepackage{graphicx}
\usepackage{wrapfig}
\usepackage{adjustbox}

%% file: format/macros.tex
\algdef{SE}[SUBALG]{Indent}{EndIndent}{}{\algorithmicend\ }%
\algtext*{Indent}
\algtext*{EndIndent}

\algrenewcommand{\algorithmiccomment}[1]{\hfill$\triangleright$ \textit{#1}}
\algnewcommand{\LongComment}[1]{\hfill$\triangleright$ {\color{RoyalBlue}\begin{minipage}[t]{\eqboxwidth{COMMENT\thealgorithm}}#1\strut\end{minipage}}}
\algdef{SE}[DOWHILE]{Do}{DoWhile}{\algorithmicdo}[1]{\algorithmicwhile\ #1}

\definecolor{mereq}{HTML}{BA160C}
\definecolor{mereqnp}{HTML}{F88379}
\definecolor{maxent}{HTML}{006D6F}
\definecolor{maxentft}{HTML}{3AB09E}
\definecolor{myexpert}{rgb}{0.188, 0.188, 0.188}
\definecolor{myprior}{rgb}{0.563, 0.563, 0.563}
\definecolor{hgdagger}{HTML}{2B4593}
\definecolor{iwr}{HTML}{468FEA}

%% file: notation_macros.tex
\usepackage{booktabs}

\usepackage{amsmath,amssymb,stmaryrd,mathtools}
\usepackage{xcolor}
\usepackage{xspace}
\usepackage{bbm}
\usepackage{nicefrac}

\definecolor{orange(sae/ece)}{rgb}{1.0, 0.49, 0.0}
\definecolor{teal(sae/ece)}{rgb}{0, 0.47, 0.52}
\definecolor{purple}{rgb}{0.74, 0.65, 1.0}
\definecolor{light_gray}{rgb}{0.9, 0.9, 0.9}
\definecolor{medium_gray}{rgb}{0.6, 0.6, 0.6} 
\definecolor{dark_gray}{rgb}{0.2, 0.2, 0.2} 
\definecolor{dark_blue}{rgb}{0.098, 0.239, 0.52}
\definecolor{dark_brown}{rgb}{0.3255, 0.004, 0.001}
\definecolor{dark_purple}{rgb}{0.478, 0.1569, 0.4863}
\definecolor{light_blue}{rgb}{0.33, 0.80, 1}

\usepackage{amsmath}

%% file: appendix.tex
\clearpage
\appendix 
\numberwithin{equation}{section}
\section{Detailed Environment Settings}
\label{app:environment_settings}
\textbf{Tasks.} We design a series of both simulated and real-world tasks featuring discrete and continuous action spaces to evaluate the effectiveness of \textsc{MEReQ}. These tasks are categorized into two experiment settings: 1) Learning from synthesized expert with heuristic-based intervention rules, and 2) human-in-the-loop (HITL) experiments.

\subsection{Learning from Synthesized Expert with Heuristic-based Intervention}
\label{app:gt_expert_policy_experiments}
In order to directly evaluate the sub-optimality of the learned policy through \textsc{MEReQ}, we specify a residual reward function and train an expert policy using this residual reward function and the prior reward function. We then define a heuristic-based intervention rule to decide when the expert should intervene or disengage. In this experiment setting, we consider two simulation environments for the highway driving task and the robot manipulation task.

\subsubsection{Highway-Sim}
\label{app:highway_sim}

\begin{figure}[b]
    \centering
    \begin{subfigure}[b]{0.16\linewidth}
        \centering
        \includegraphics[width=\linewidth]{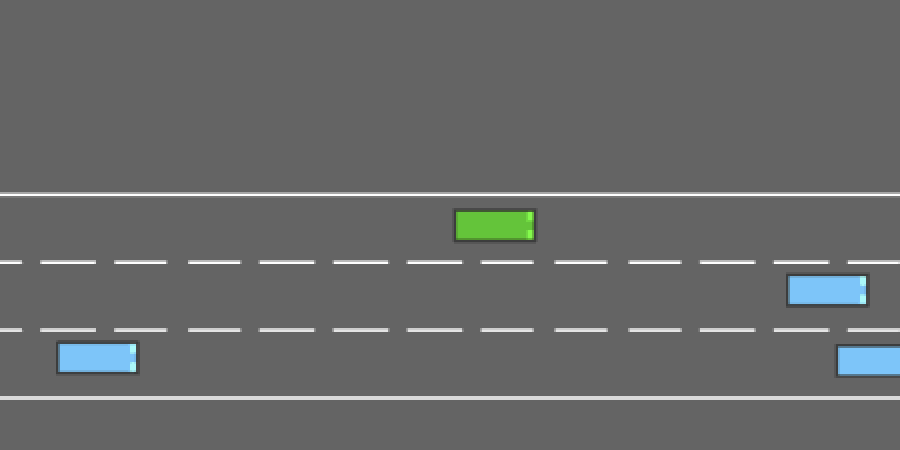}
    \end{subfigure}\hfill
    \begin{subfigure}[b]{0.16\linewidth}
        \centering
        \includegraphics[width=\linewidth]{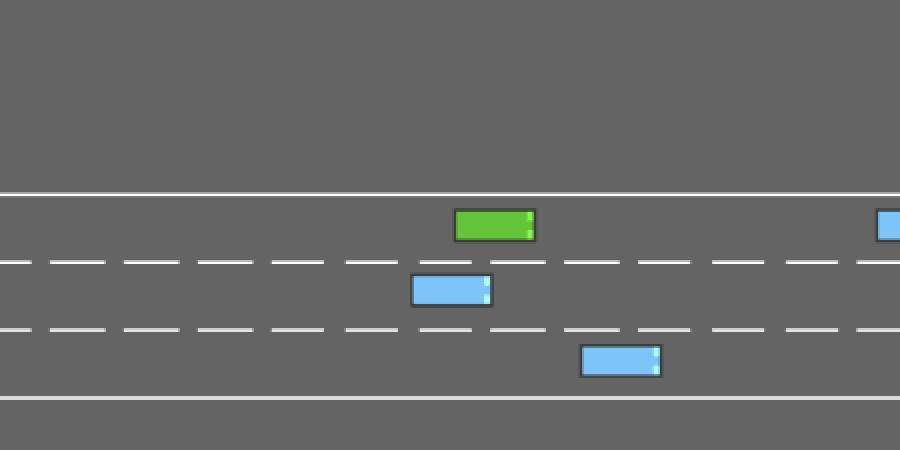}
    \end{subfigure}\hfill
    \begin{subfigure}[b]{0.16\linewidth}
        \centering
        \includegraphics[width=\linewidth]{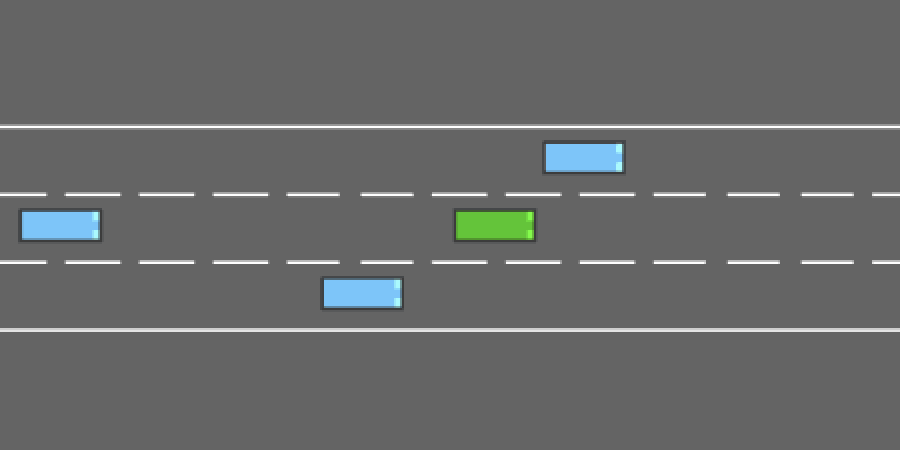}
    \end{subfigure}\hfill
    \begin{subfigure}[b]{0.16\linewidth}
        \centering
        \includegraphics[width=\linewidth]{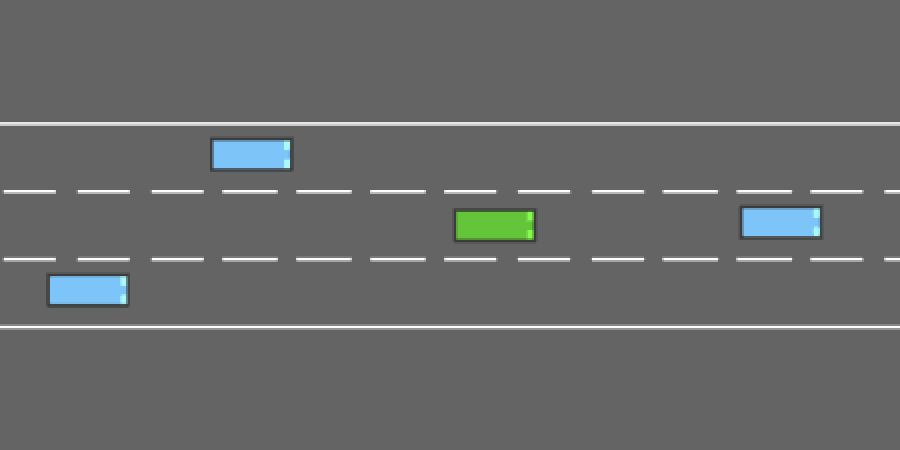}
    \end{subfigure}\hfill
    \begin{subfigure}[b]{0.16\linewidth}
        \centering
        \includegraphics[width=\linewidth]{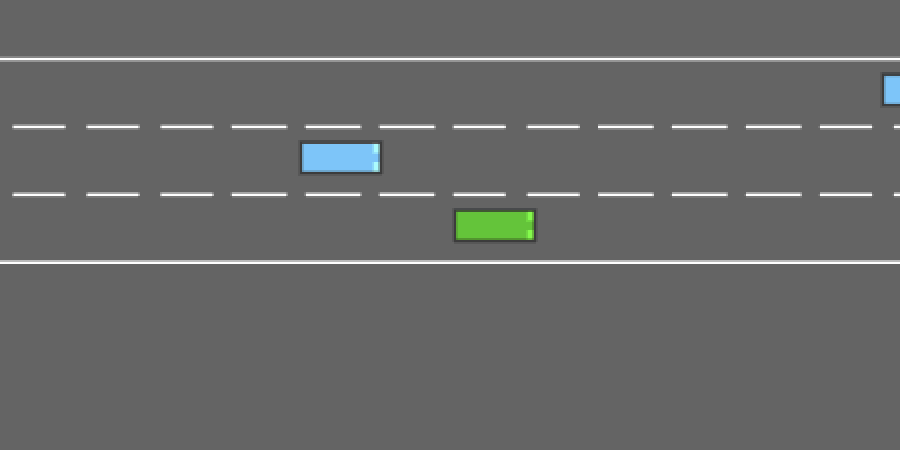}
    \end{subfigure}\hfill
    \begin{subfigure}[b]{0.16\linewidth}
        \centering
        \includegraphics[width=\linewidth]{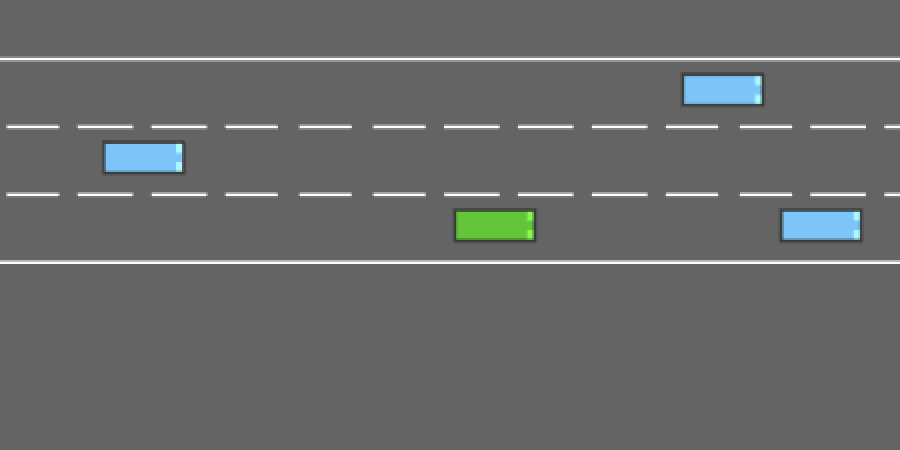}
    \end{subfigure}
    \caption{\textbf{\textit{Highway-Sim} Sample Roll-out.} The green box is the ego vehicle, and the blue boxes are the surrounding vehicles. The bird-eye-view bounding box follows the ego vehicle.}
    \label{fig:highway_sim_demo}
\end{figure}

\textbf{Overview.} We adopt the \texttt{highway-env}~\citep{highway-env} environment for this task. The ego vehicle must navigate traffic using discrete actions to control speed and change lanes. The expert policy prefers the ego vehicle to stay in the right-most lane of a three-lane highway. Expert intervention is based on KL divergence between the expert and learned policies: the expert steps in if there is a significant mismatch for several consecutive steps and disengages once the distributions align for a sufficient number of steps. Each episode lasts for 40 steps. The sample roll-out is shown in Fig.~\ref{fig:highway_sim_demo}.

\textbf{Rewards Design.} In \textit{Highway-Sim} there are 5 available discrete actions for controlling the ego vehicle: $\mathcal{A}=\{\mathbf{a}_\texttt{lane\_left},\mathbf{a}_\texttt{idle},\mathbf{a}_\texttt{lane\_right},\mathbf{a}_\texttt{faster},\mathbf{a}_\texttt{slower}\}$. Rewards are based on 3 features: $\textbf{f}=\{\mathbf{f}_\texttt{collision},\mathbf{f}_\texttt{high\_speed},\mathbf{f}_\texttt{right\_lane}\}$, defined as follows:
\begin{itemize}
    \item $\mathbf{f}_\texttt{collision}\in\{0,1\}$: $0$ indicates no collision, $1$ indicates a collision with a vehicle.
    \item $\mathbf{f}_\texttt{high\_speed}\in[0,1]$: This feature is $1$ when the ego vehicle's speed exceeds $30$ m/s, and linearly decreases to  $0$ for speeds down to $20$ m/s.
    \item $\mathbf{f}_\texttt{right\_lane}\in\{0,0.5,1\}$: This feature is $1$ for the right-most lane, $0.5$ for the middle lane, and $0$ for the left-most lane.
\end{itemize}

The reward is defined as a linear combination of the feature set with the weights $\theta$. For the prior policy, we define the basic reward as
\begin{equation}
    r = -0.5~*~\mathbf{f}_\texttt{collision}+0.4~*~\mathbf{f}_\texttt{high\_speed}.
\end{equation}
For the expert policy, we define its reward as the basic reward with an additional term on $\mathbf{f}_\texttt{right\_lane}$
\begin{equation}
\begin{aligned}
    r_\text{expert} &= -0.5~*~\mathbf{f}_\texttt{collision}+0.4~*~\mathbf{f}_\texttt{high\_speed}\\
    &\quad+ 0.5~*~\mathbf{f}_\texttt{right\_lane}.
\end{aligned}
\end{equation}
Both prior and expert policy are trained using Deep Q-Network (DQN)~\citep{mnih2013playing} with the reward defined above in Gymnasium~\citep{brockman2016openai} environment. The hyperparameters are shown in Tab.~\ref{tab:DQN_hyper}.

\textbf{Intervention Rule.}
The expert intervention is determined by the KL divergence between the expert policy $\pi_\mathrm{e}$ and the learner policy $\hat{\pi}$ given the same state observation $\mathbf{s}$, denoted as $D_\mathrm{KL}(\hat{\pi}(\mathbf{a}|\mathbf{s})\parallel\pi_\mathrm{e}(\mathbf{a}|\mathbf{s}))$. At each time step, the state observation is fed into both policies to obtain the expert action $\mathbf{a}_\mathrm{e}$, the learner action $\hat{\mathbf{a}}$, and the expert action distribution $\pi_\mathrm{e}(\mathbf{a}|\mathbf{s})$, defined as
\begin{equation}
    \pi_\mathrm{e}(\mathbf{a}|\mathbf{s}) = \frac{\exp(Q^\star_\mathrm{e}(\mathbf{s},\mathbf{a}))}{\sum \exp(Q^\star_\mathrm{e}(\mathbf{s},a_i))},
\end{equation}
where $Q^\star_\mathrm{e}$ is the soft $Q$-function. The learner's policy distribution $\hat{\pi}(\mathbf{a}|\mathbf{s})$ is treated as a \textit{delta distribution} of the learner action $\delta[\mathbf{a}_\mathrm{l}]$.

We define heuristic thresholds $(D_\mathrm{KL,upper},D_\mathrm{KL,lower})=(1.62,1.52)$. If the learner policy is in control and $D_\mathrm{KL}\geq D_\mathrm{KL,upper}$ for 2 consecutive steps, the expert policy takes over; During expert control, if $D_\mathrm{KL}\leq D_\mathrm{KL,lower}$ for 4 consecutive steps, the expert disengages. Each expert intervention must last at least 4 steps.
 
\subsubsection{Bottle-Pushing-Sim}
\label{app:bottle_pushing_sim}

\begin{figure}[t]
    \centering
    \begin{subfigure}[b]{0.24\linewidth}
        \centering
        \includegraphics[width=\linewidth]{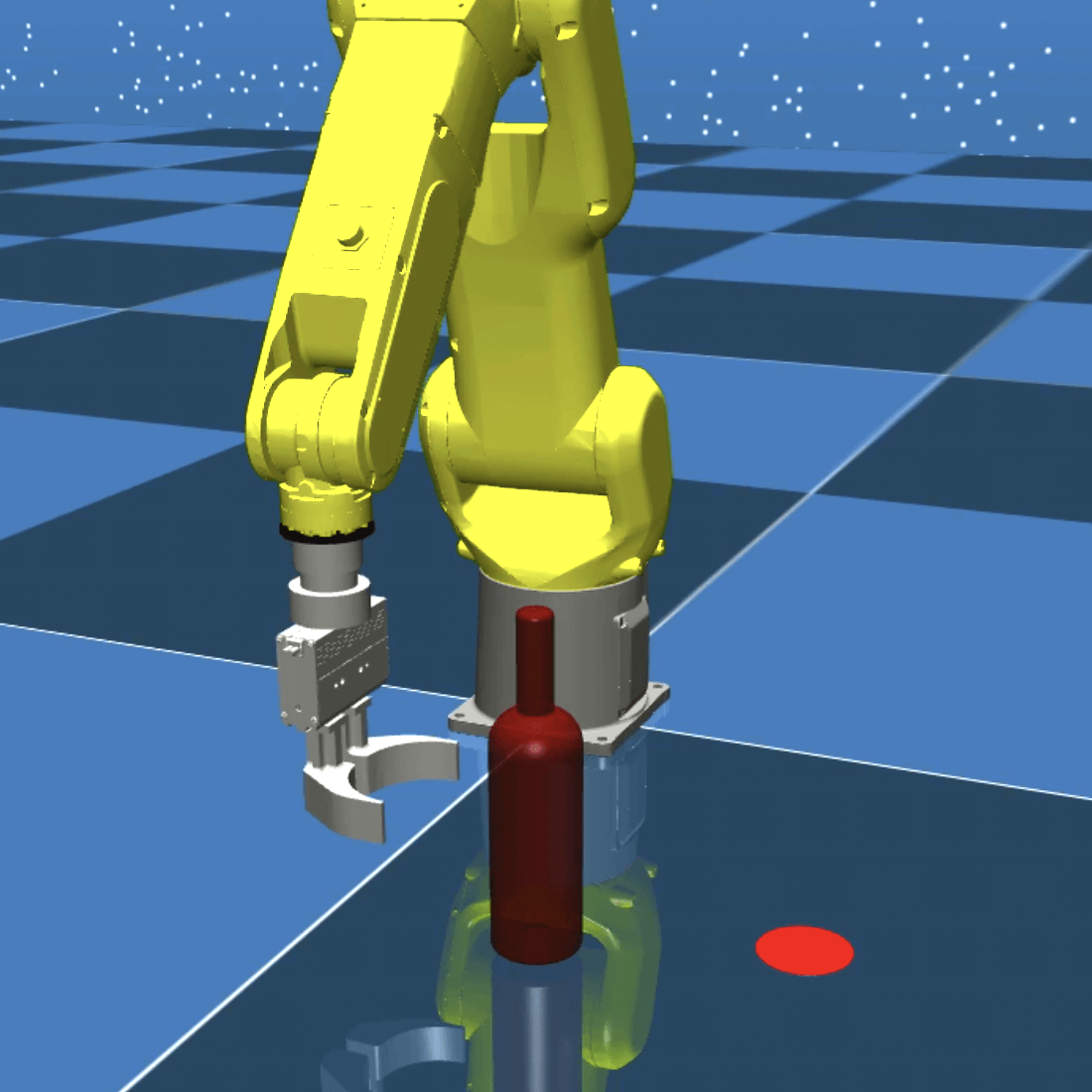}
    \end{subfigure}\hfill
    \begin{subfigure}[b]{0.24\linewidth}
        \centering
        \includegraphics[width=\linewidth]{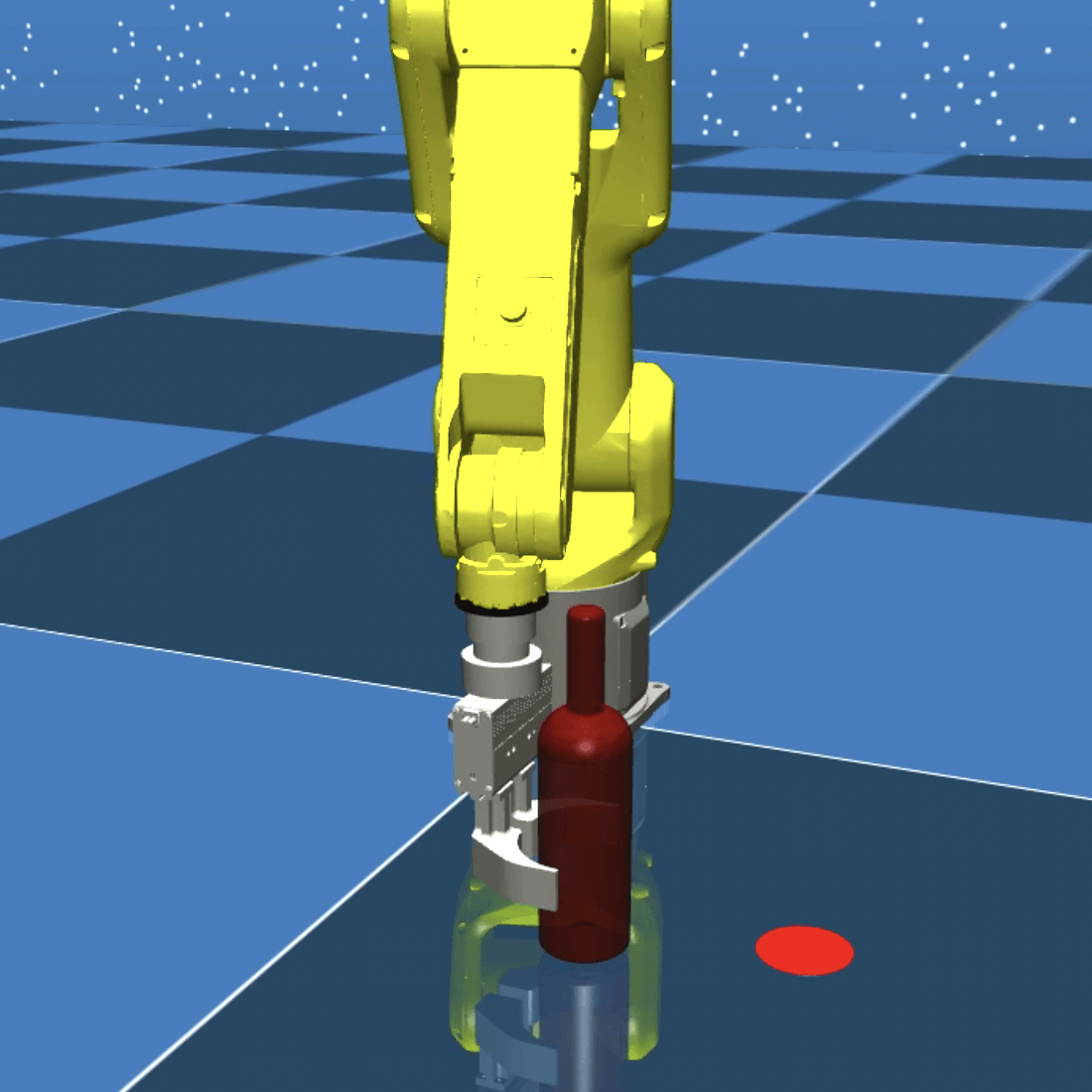}
    \end{subfigure}\hfill
    \begin{subfigure}[b]{0.24\linewidth}
        \centering
        \includegraphics[width=\linewidth]{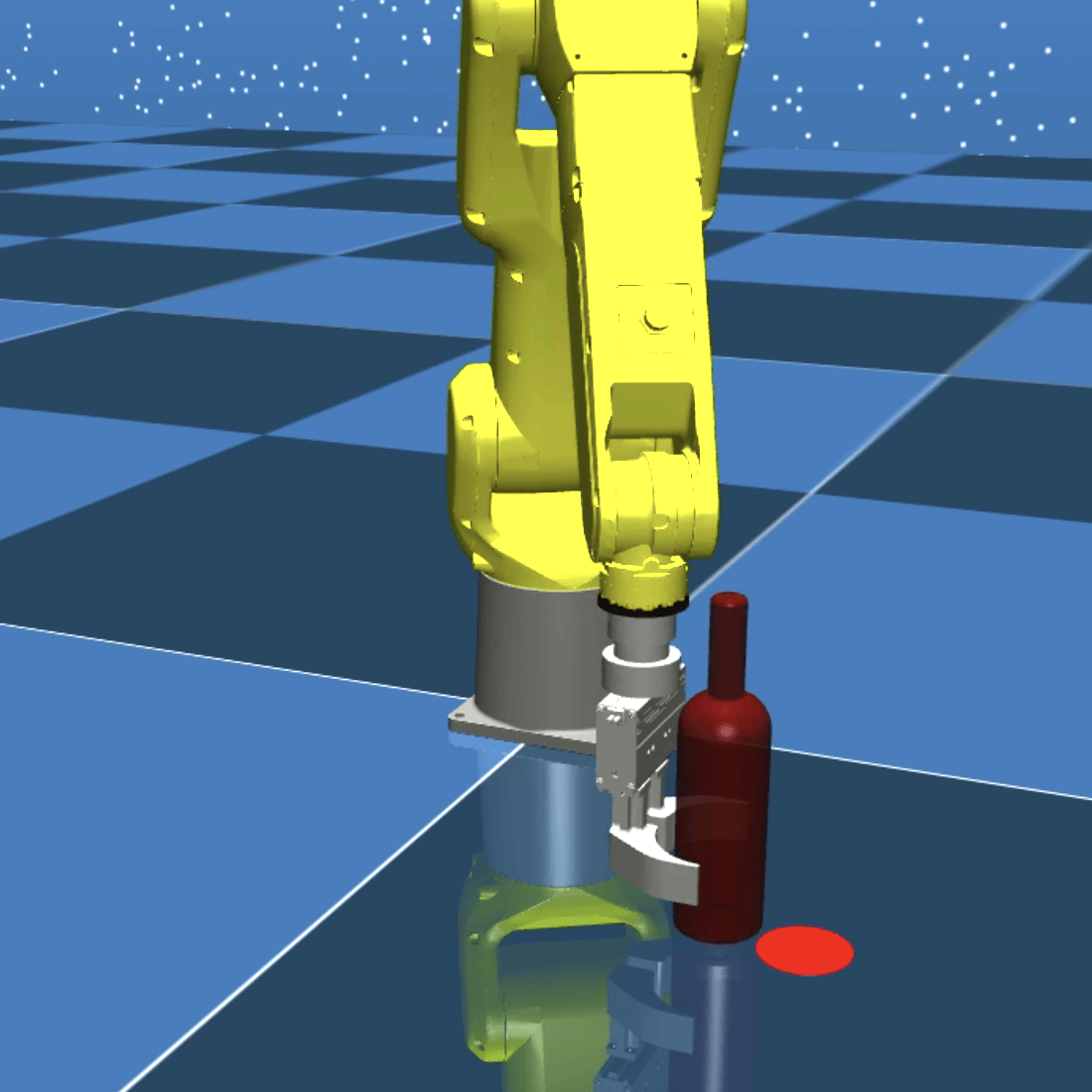}
    \end{subfigure}\hfill
    \begin{subfigure}[b]{0.24\linewidth}
        \centering
        \includegraphics[width=\linewidth]{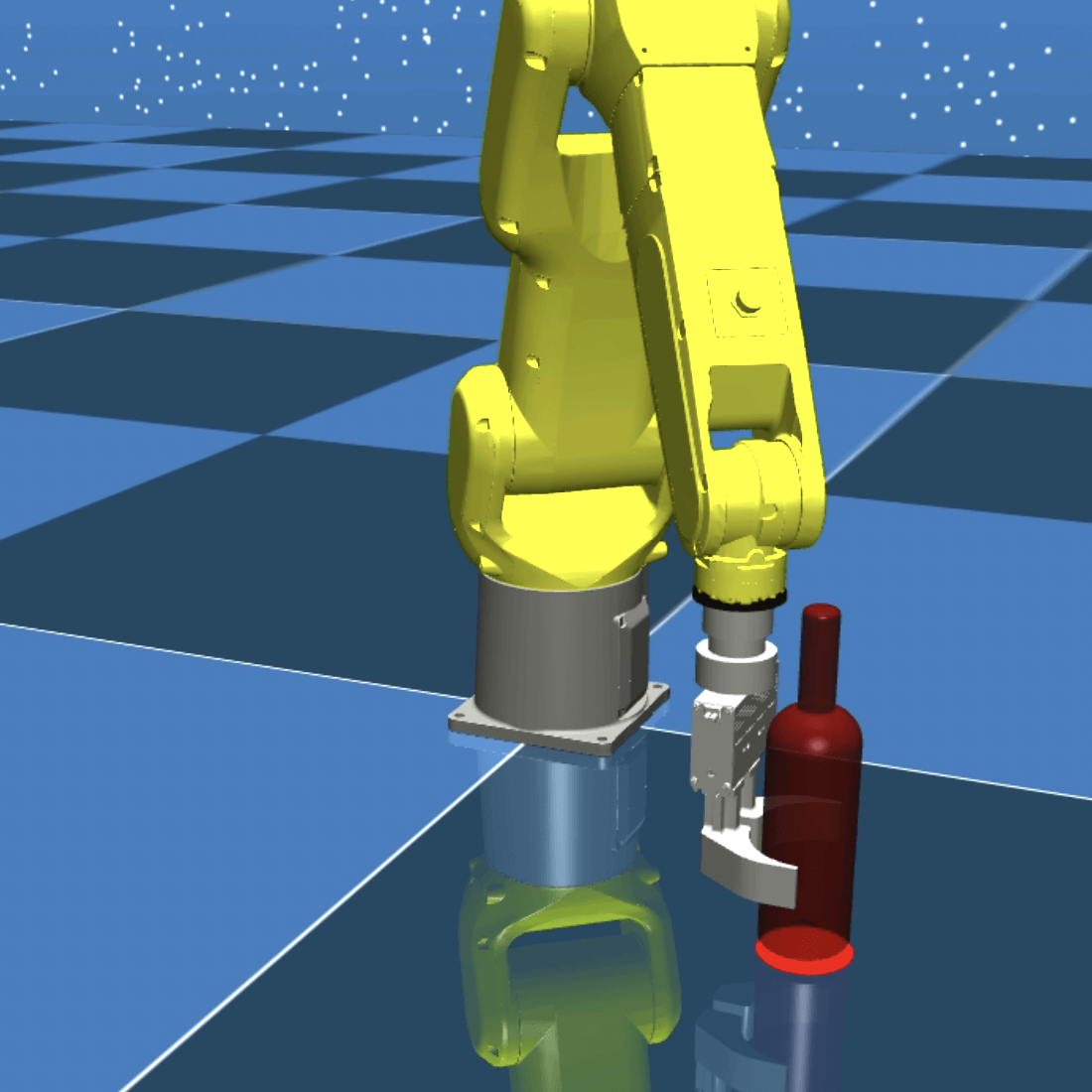}
    \end{subfigure}
    \caption{\textbf{\textit{Bottle-Pushing-Sim} Sample Roll-out.} The location of the wine bottle and the goal are randomly initialized for each episode.}
    \label{fig:bottle_pushing_sim}
\end{figure}

\textbf{Overview.} A 6-DoF robot arm is tasked with pushing a wine bottle to a random goal position. The expert policy prefers pushing from the bottom for safety. Expert intervention is based on state observation: the expert engages if the tooltip is too high, risking the bottle tilting for several consecutive steps, and disengages when the tooltip stays low enough for a sufficient number of steps. Each episode lasts for 100 steps. The sample roll-out is shown in Fig.~\ref{fig:bottle_pushing_sim}.

\textbf{Rewards Design.}
In \textit{Bottle-Pushing-Sim}, the action space $\mathbf{a}\in\mathbb{R}^3$ is continuous, representing end-effector movements along the global $x$, $y$, and $z$ axes. Each dimension ranges from $-1$ to $1$, with positive values indicating movement in the positive direction and negative values indicating movement in the negative direction along the respective axes. All values are in centimeter. The rewards are based on 4 features: $\mathbf{f}=\{\mathbf{f}_\texttt{tip2bottle},\mathbf{f}_\texttt{bottle2goal},\mathbf{f}_\texttt{control\_effort},\mathbf{f}_\texttt{table\_distance}\}$, with:
\begin{itemize}
    \item $\mathbf{f}_\texttt{tip2bottle}\in[0,1]$: This feature is $1$ when the distance between the end-effector tool tip and the wine bottle's geometric center exceeds 30 cm, and decreases linearly to $0$ as the distance approaches $0$ cm.
    \item $\mathbf{f}_\texttt{bottle2goal}\in[0,1]$: This feature is $1$ when the distance between the wine bottle and the goal exceeds $30$ cm, and decreases linearly to $0$ as the distance approaches $0$ cm.
    \item $\mathbf{f}_\texttt{control\_effort}\in[0,1]$: This feature is $1$ when the end-effector acceleration exceeds $5\times 10^{-3}$ m/s$^2$, and decreases linearly to $2$ as the acceleration approaches $0$.
    \item $\mathbf{f}_\texttt{table\_distance}\in[0,1]$: This feature is $1$ when the distance between the end-effector tool tip and the table exceeds 10 cm, and decreases linearly to $0$ as the distance approaches $0$ cm.
\end{itemize}


The reward is defined as a linear combination of the feature set with the weights $\theta$. For the prior policy, we define the basic reward as
\begin{equation}
\begin{aligned}
    r &= -1.0~*~\mathbf{f}_\texttt{tip2bottle}-1.0~*~\mathbf{f}_\texttt{bottle2goal}\\
    &\quad-0.2~*~\mathbf{f}_\texttt{control\_effort}.
\end{aligned}
\end{equation}
For the expert policy, we define the expert reward as the basic reward with an additional term on $\mathbf{f}_\texttt{table\_distance}$
\begin{equation}
\begin{aligned}
    r_\text{expert} &= -1.0~*~\mathbf{f}_\texttt{tip2bottle}-1.0~*~\mathbf{f}_\texttt{bottle2goal}\\
    &\quad-0.2~*~\mathbf{f}_\texttt{control\_effort}-0.8~*~\mathbf{f}_\texttt{table\_distance}.
\end{aligned}
\end{equation}

Both prior and expert policy are trained using Soft Actor-Critic (SAC)~\citep{haarnoja2018soft} with the rewards defined above in MuJoCo~\citep{todorov2012mujoco} environment. The hyperparameters are shown in Tab.~\ref{tab:SAC_hyper}.

\textbf{Intervention Rule.}
During learner policy execution, the expert policy takes over if either of the following conditions is met for $5$ consecutive steps:

\begin{enumerate}
    \item After $20$ time steps, the bottle is not close to the goal (\(\mathbf{f}_\texttt{bottle2goal} \geq 3\) cm) and the distance between the end-effector and the table exceeds $3$ cm (\(\mathbf{f}_\texttt{table\_distance} \geq 3\) cm).
    \item After $40$ time steps, the bottle is not close to the goal (\(\mathbf{f}_\texttt{bottle2goal} \geq 3\) cm) and the bottle movement in the past time step is less than $0.1$ cm.
\end{enumerate}

During expert control, the expert disengages if either of the following conditions is met for $3$ consecutive steps:

\begin{enumerate}
    \item The distance between the end-effector and the table exceeds $3$ cm (\(\mathbf{f}_\texttt{table\_distance} \leq 3\) cm) and the bottle movement in the past time step is greater than $0.1$ cm.
    \item The bottle is close to the goal (\(\mathbf{f}_\texttt{bottle2goal} \leq 3\) cm).
\end{enumerate}

\subsubsection{Erasing-Sim}
\label{app:erasing_sim}

\begin{figure}[t]
    \centering
    \begin{subfigure}[b]{0.24\linewidth}
        \centering
        \includegraphics[width=\linewidth]{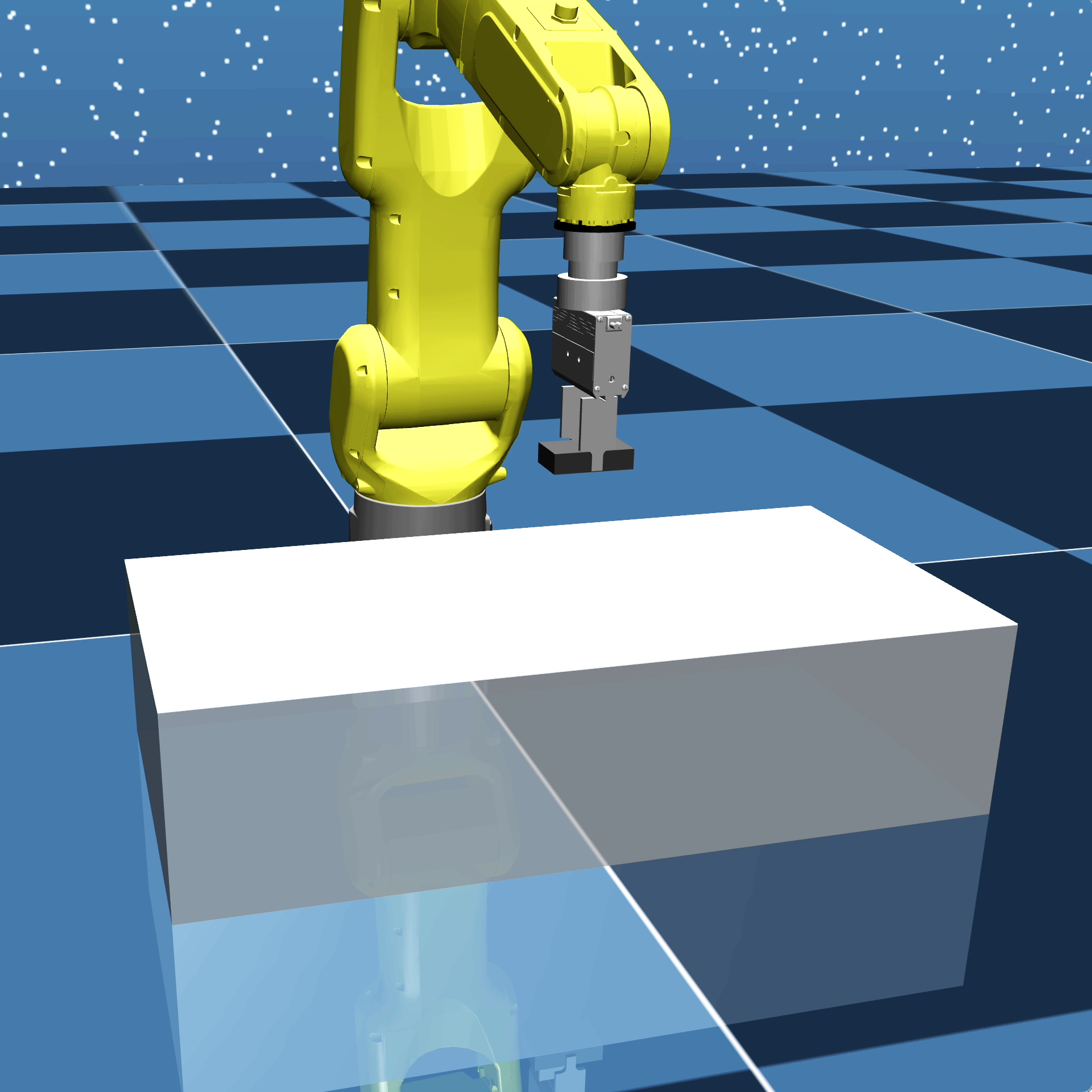}
    \end{subfigure}\hfill
    \begin{subfigure}[b]{0.24\linewidth}
        \centering
        \includegraphics[width=\linewidth]{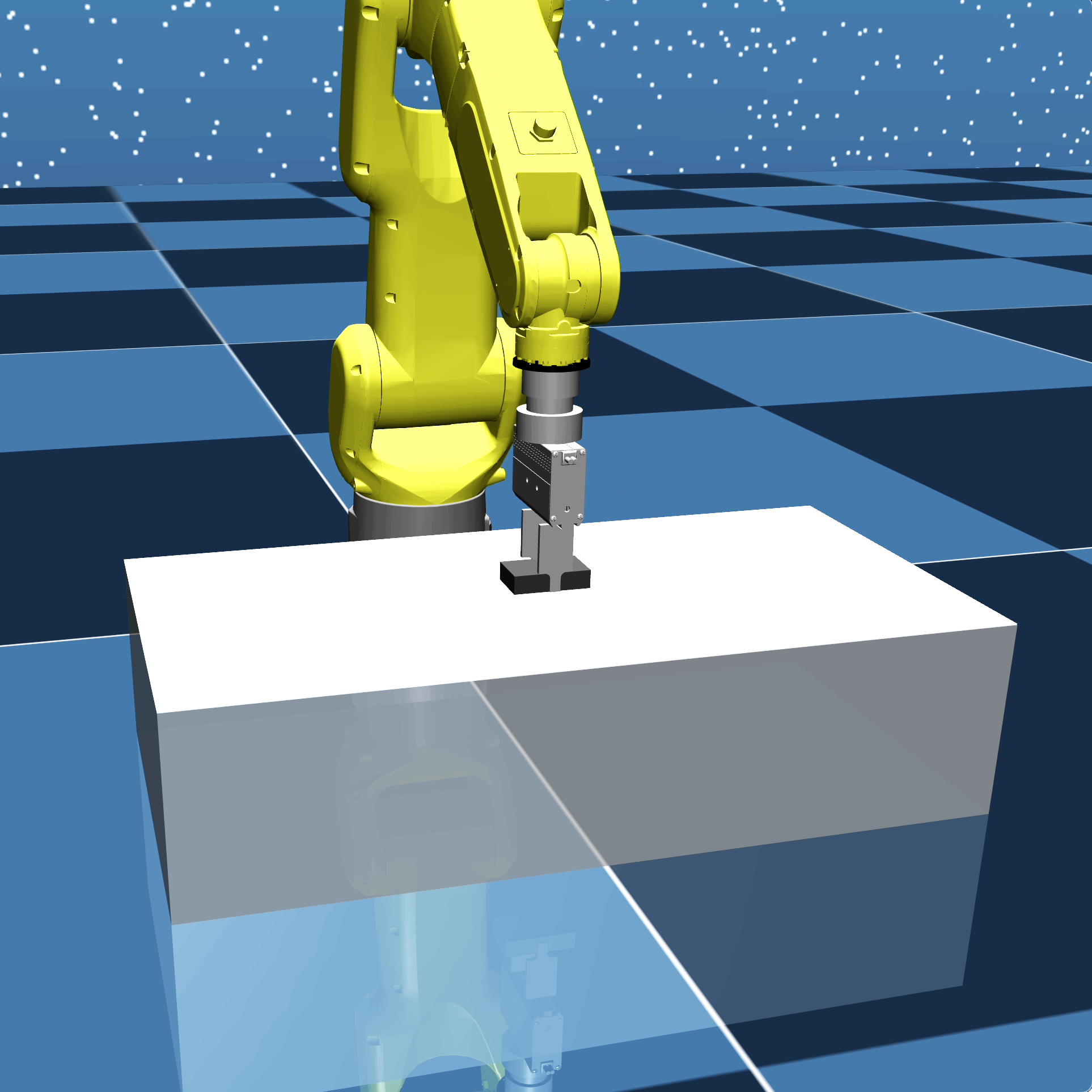}
    \end{subfigure}\hfill
    \begin{subfigure}[b]{0.24\linewidth}
        \centering
        \includegraphics[width=\linewidth]{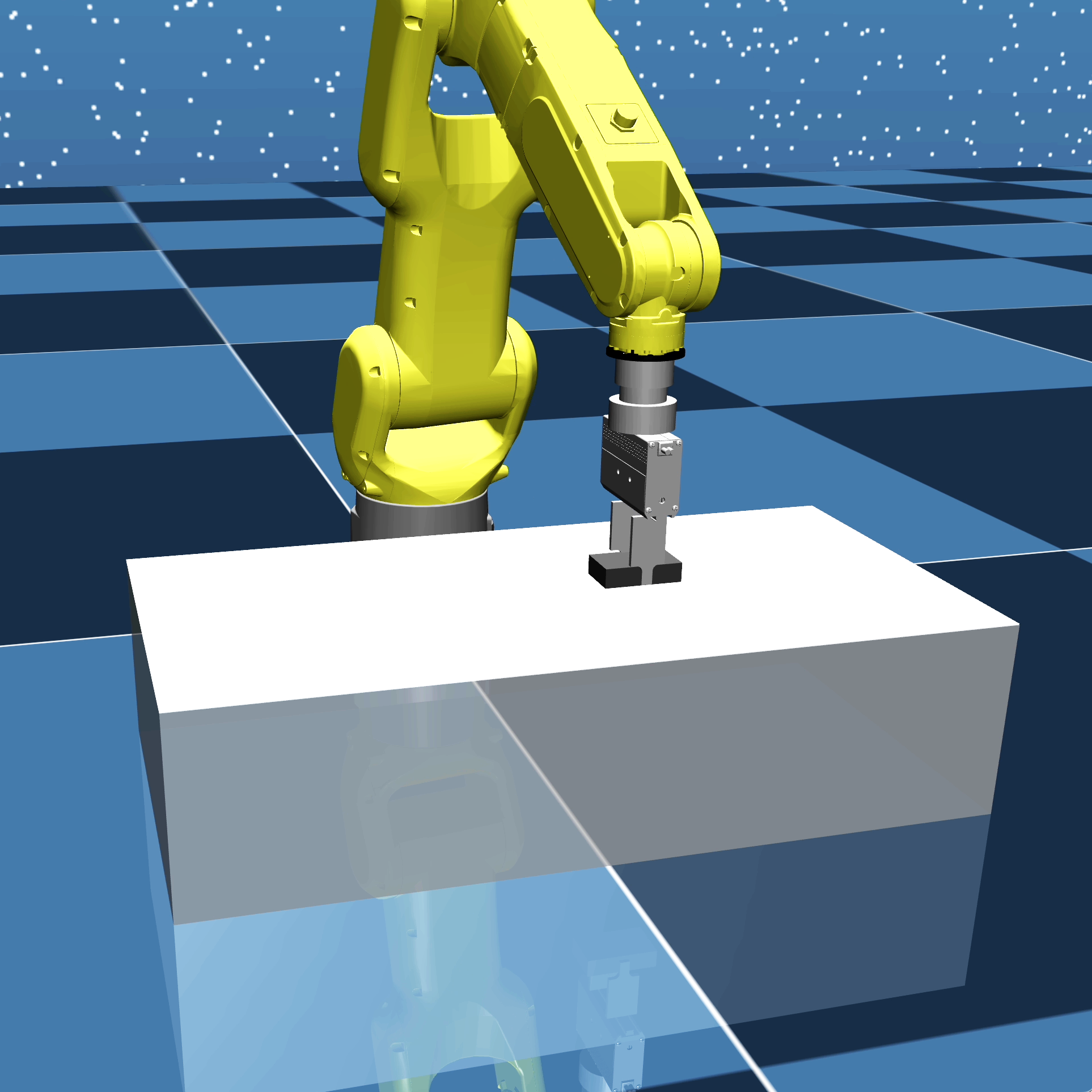}
    \end{subfigure}\hfill
    \begin{subfigure}[b]{0.24\linewidth}
        \centering
        \includegraphics[width=\linewidth]{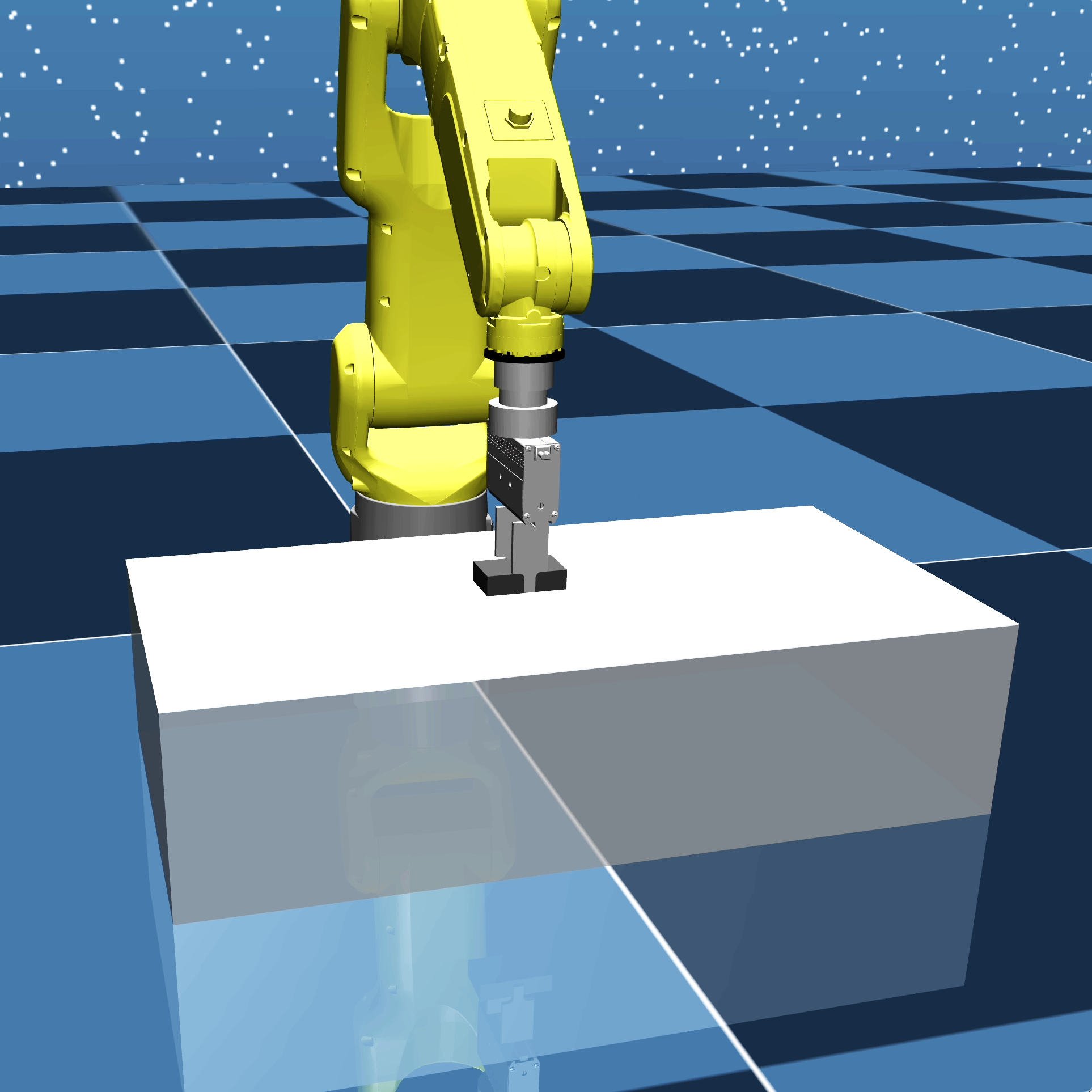}
    \end{subfigure}
    \caption{\textbf{\textit{Erasing-Sim} Sample Roll-out.} The location of the whiteboard is fixed for each episode.}
    \label{fig:erasing_sim}
\end{figure}

\textbf{Overview.} A 6-DoF robot arm is tasked with erasing marker on a whiteboard on the table with an eraser. The expert policy prefers applying a larger normal force to ensure the erasing performance. Expert intervention is based on the contact force of the end-effector: the expert engages if the normal force applied by the end-effector is too small for several consecutive steps, and disengages after a fixed number of steps. Each episode lasts for 100 steps. The sample roll-out is shown in Fig.~\ref{fig:erasing_sim}.

\textbf{Rewards Design.}
In \textit{Erasing-Sim}, the action space $\mathbf{a}\in\mathbb{R}^3$ is continuous, representing end-effector movements along the global $x$, $y$, and $z$ axes. Each dimension ranges from $-1$ to $1$, with positive values indicating movement in the positive direction and negative values indicating movement in the negative direction along the respective axes. All values are in centimeter. The rewards are based on 4 features: $\mathbf{f}=\{\mathbf{f}_\texttt{tip\_hor\_move},\mathbf{f}_\texttt{tip\_ver\_dist},\mathbf{f}_\texttt{control\_effort},\mathbf{f}_\texttt{tip\_force}\}$, with:
\begin{itemize}
    \item $\mathbf{f}_\texttt{tip\_hor\_move}\in[0,1]$: This feature is $1$ when the horizontal movement of the end-effector since last step exceeds 0.6 cm, and decreases linearly to $0$ as the movement approaches $0$.
    \item $\mathbf{f}_\texttt{tip\_ver\_dist}\in[0,1]$: This feature is $1$ when the distance between the eraser and the whiteboard exceeds $4$ cm, and decreases linearly to $0$ as the distance approaches $0$ cm.
    \item $\mathbf{f}_\texttt{control\_effort}\in[0,1]$: This feature is $1$ when the end-effector acceleration exceeds $5\times 10^{-3}$ m/s$^2$, and decreases linearly to $2$ as the acceleration approaches $0$.
    \item $\mathbf{f}_\texttt{tip\_force}\in[0,1]$: This feature is $1$ when the normal force applied by the eraser exceeds $4$ N, and decreases linearly to $0$ as the normal force approaches $0$ N.
\end{itemize}

The reward is defined as a linear combination of the feature set with the weights $\theta$. For the prior policy, we define the basic reward as
\begin{equation}
\begin{aligned}
    r &= 1.0~*~\mathbf{f}_\texttt{tip\_hor\_move}-1.0~*~\mathbf{f}_\texttt{tip\_ver\_dist}\\
    &\quad-0.2~*~\mathbf{f}_\texttt{control\_effort}.
\end{aligned}
\end{equation}
For the expert policy, we define the expert reward as the basic reward with an additional term on $\mathbf{f}_\texttt{table\_distance}$
\begin{equation}
\begin{aligned}
    r_\text{expert} &= 1.0~*~\mathbf{f}_\texttt{tip\_hor\_move}-1.0~*~\mathbf{f}_\texttt{tip\_ver\_dist}\\
    &\quad-0.2~*~\mathbf{f}_\texttt{control\_effort}+2.0~*~\mathbf{f}_\texttt{tip\_force}.
\end{aligned}
\end{equation}

Both prior and expert policy are trained using Soft Actor-Critic (SAC)~\citep{haarnoja2018soft} with the rewards defined above in MuJoCo~\citep{todorov2012mujoco} environment. The hyperparameters are shown in Tab.~\ref{tab:SAC_hyper}.

\textbf{Intervention Rule.}
During learner policy execution, the expert policy takes over if the normal force applied by the end-effector is smaller than 2 N ($\mathbf{f}_\texttt{tip\_force} < 2\text{ N}$) for 5 consecutive steps. Expert control will last for 5 steps and automatically disengages.

\subsubsection{Pillow-Grasping-Sim}
\label{app:pillow_grasping_sim}

\textbf{Overview.} A 6-DoF robot arm is tasked with grasping a pillow with a parallel two-finger gripper. The expert policy prefers grasping from the center. Expert intervention is based on the state observation: the expert engages if the gripper is not going lower and closer to the center, and disengages when the gripper is actively moving towards the center. Each episode lasts for 100 steps. The sample roll-out is shown in Fig.~\ref{fig:erasing_sim}.

\begin{figure}[t]
    \centering
    \begin{subfigure}[b]{0.24\linewidth}
        \centering
        \includegraphics[width=\linewidth]{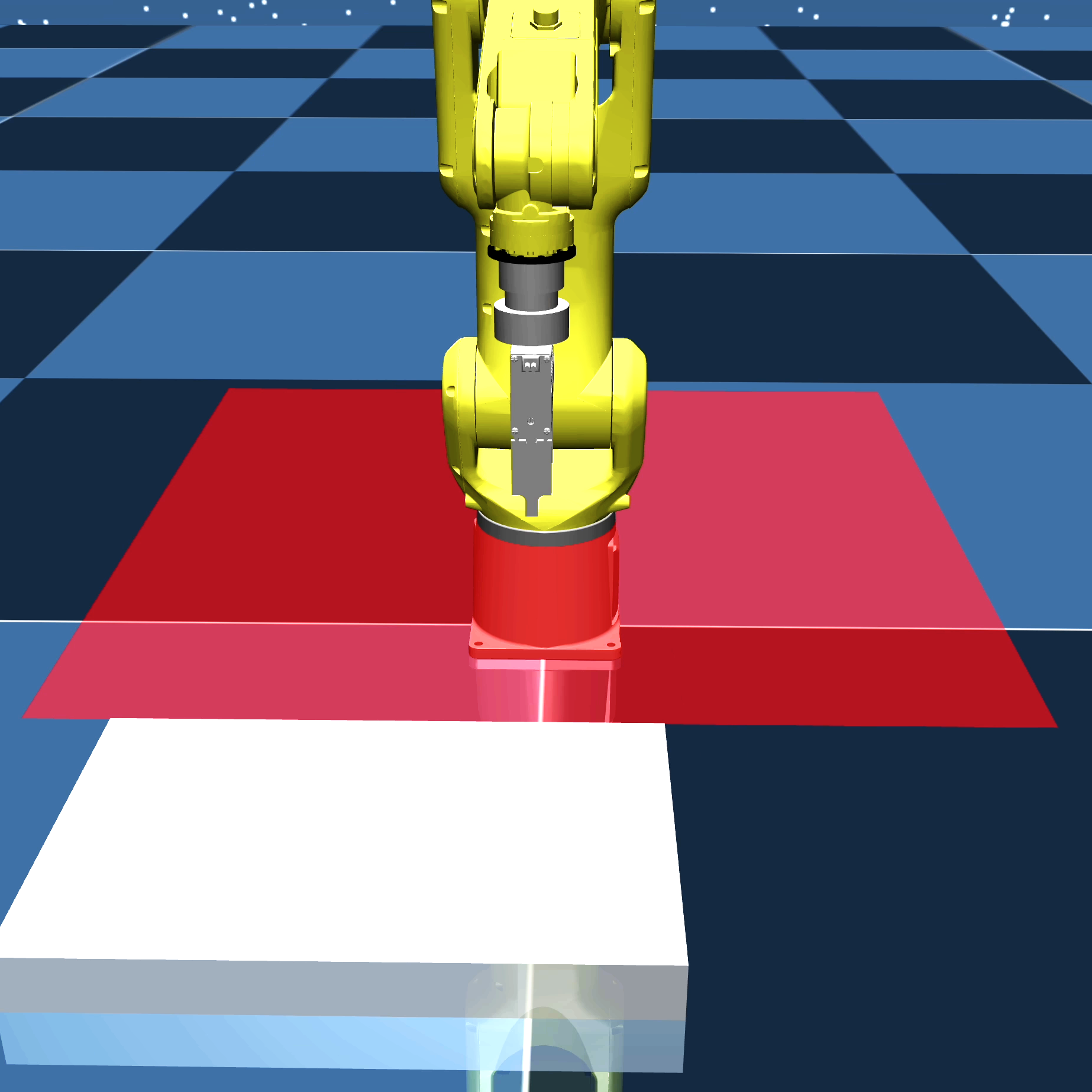}
    \end{subfigure}\hfill
    \begin{subfigure}[b]{0.24\linewidth}
        \centering
        \includegraphics[width=\linewidth]{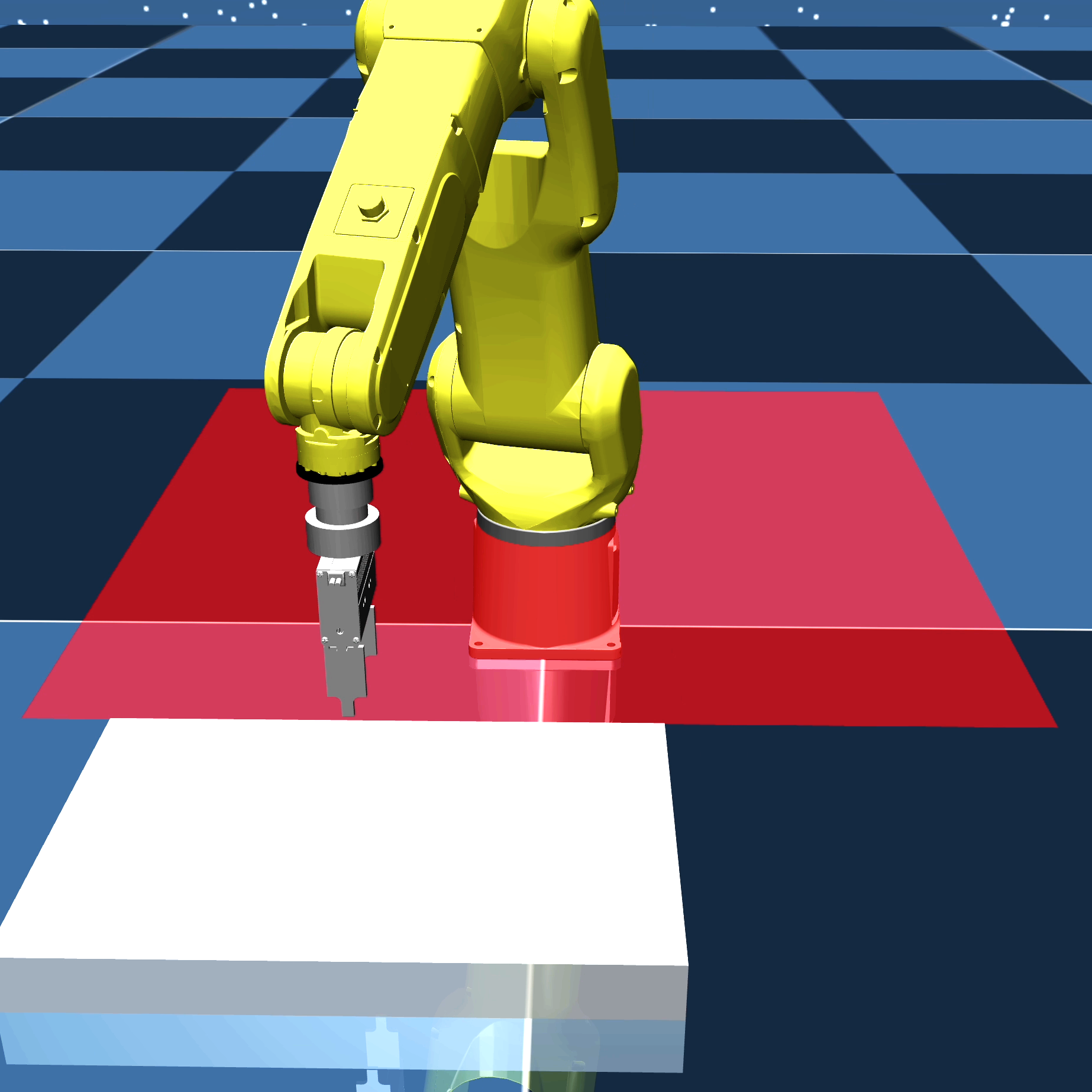}
    \end{subfigure}\hfill
    \begin{subfigure}[b]{0.24\linewidth}
        \centering
        \includegraphics[width=\linewidth]{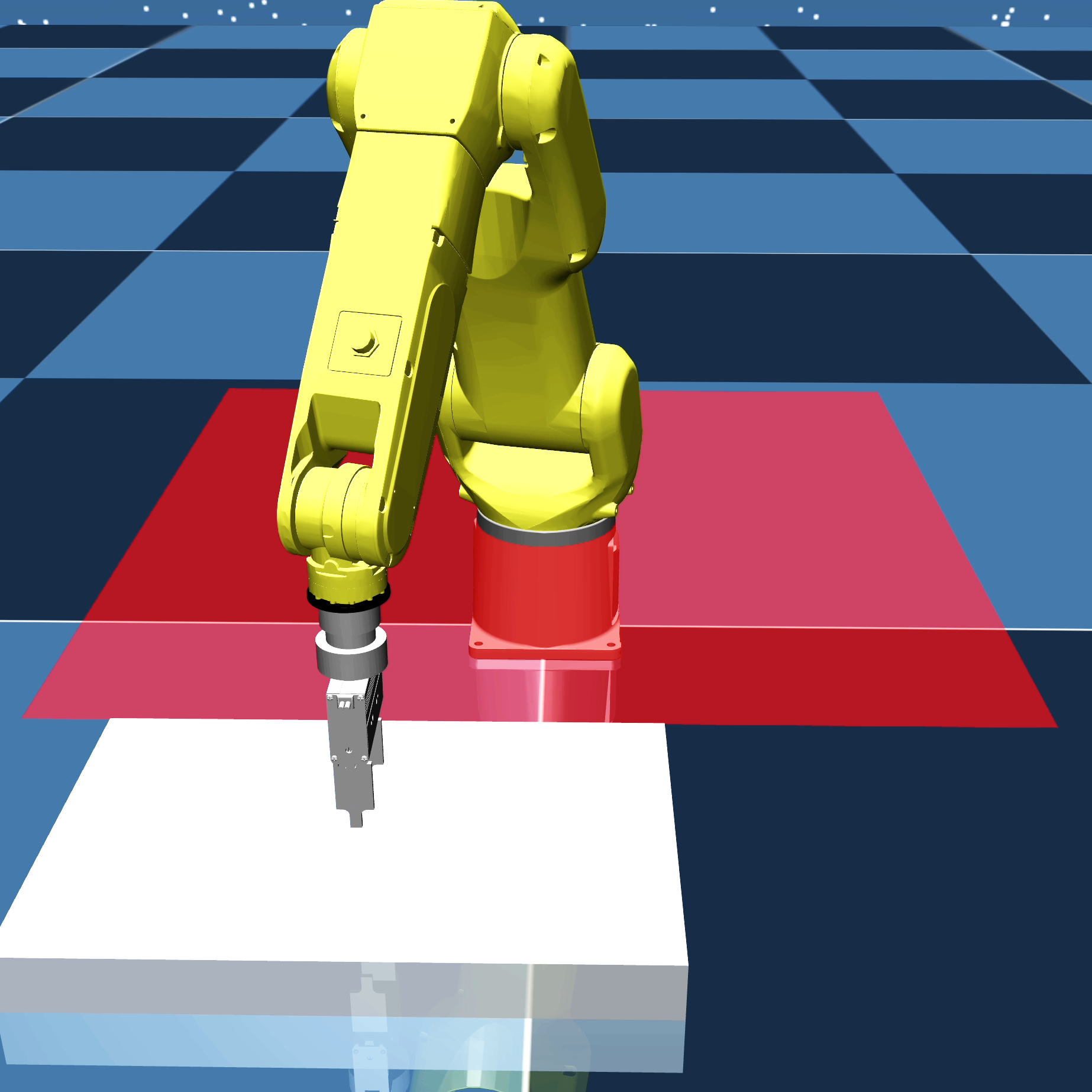}
    \end{subfigure}\hfill
    \begin{subfigure}[b]{0.24\linewidth}
        \centering
        \includegraphics[width=\linewidth]{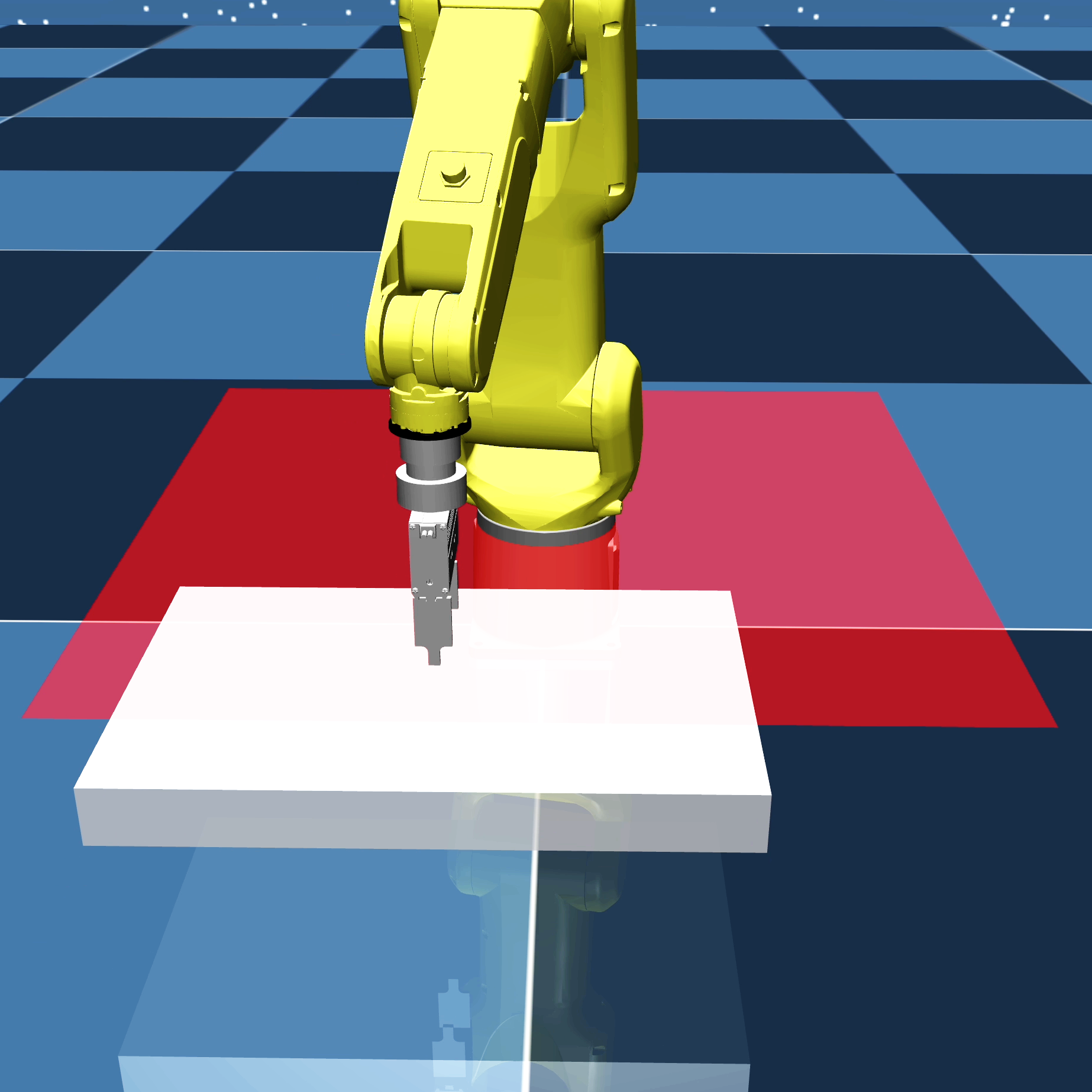}
    \end{subfigure}
    \caption{\textbf{\textit{Pillow-Grasping-Sim} Sample Roll-out.} The expert prefers grasping from the center for improved success rate.}
    \label{fig:pillow_grasping_sim}
\end{figure}

\textbf{Rewards Design.}
In \textit{Erasing-Sim}, the action space $\mathbf{a}\in\mathbb{R}^3$ is continuous, representing end-effector movements along the global $x$, $y$, and $z$ axes. Each dimension ranges from $-1$ to $1$, with positive values indicating movement in the positive direction and negative values indicating movement in the negative direction along the respective axes. All values are in centimeter. The gripper will automatically close once the end-effector reach the pillow. The rewards are based on 4 features: $\mathbf{f}=\{\mathbf{f}_\texttt{tip2pillow},\mathbf{f}_\texttt{pillow\_height},\mathbf{f}_\texttt{control\_effort},\mathbf{f}_\texttt{tip2center}\}$, with:
\begin{itemize}
    \item $\mathbf{f}_\texttt{tip2pillow}\in[0,1]$: This feature is $1$ when the vertical movement of the end-effector towards the surface of the pillow since last step exceeds $0.5$ cm, and decreases linearly to $0$ as the end-effector moving away from the surface of the pillow for more than $0.5$ cm.
    \item $\mathbf{f}_\texttt{pillow\_height}\in[0,1]$: This feature is $1$ when the distance between the pillow and the table surface $5$ cm, and decreases linearly to $0$ as the distance approaches $0$ cm.
    \item $\mathbf{f}_\texttt{control\_effort}\in[0,1]$: This feature is $1$ when the end-effector acceleration exceeds $5\times 10^{-3}$ m/s$^2$, and decreases linearly to $2$ as the acceleration approaches $0$.
    \item $\mathbf{f}_\texttt{tip2center}\in[0,1]$: This feature is $1$ when the movement of the end-effector towards the center of the pillow since last step exceeds $0.5$ cm, and decreases linearly to $0$ as the end-effector moving away from the center of the pillow for more than $0.5$ cm.
\end{itemize}

The reward is defined as a linear combination of the feature set with the weights $\theta$. For the prior policy, we define the basic reward as
\begin{equation}
\begin{aligned}
    r &= -0.5~*~\mathbf{f}_\texttt{tip2pillow}+2.0~*~\mathbf{f}_\texttt{pillow\_height}\\
    &\quad-0.2~*~\mathbf{f}_\texttt{control\_effort}.
\end{aligned}
\end{equation}
For the expert policy, we define the expert reward as the basic reward with an additional term on $\mathbf{f}_\texttt{table\_distance}$
\begin{equation}
\begin{aligned}
    r_\text{expert} &= -0.5~*~\mathbf{f}_\texttt{tip2pillow}+2.0~*~\mathbf{f}_\texttt{pillow\_height}\\
    &\quad-0.2~*~\mathbf{f}_\texttt{control\_effort}-0.8~*~\mathbf{f}_\texttt{tip2center}.
\end{aligned}
\end{equation}

Both prior and expert policy are trained using Soft Actor-Critic (SAC)~\citep{haarnoja2018soft} with the rewards defined above in MuJoCo~\citep{todorov2012mujoco} environment. The hyperparameters are shown in Tab.~\ref{tab:SAC_hyper}.

\textbf{Intervention Rule.}
During learner policy execution, the expert policy takes over if:
\begin{enumerate}
    \item The horizontal movement of the end-effector towards the center of the pillow during last step is less than a pre-defined threshold for 5 consecutive steps. The threshold varies depending on the current vertical distance between the end-effector and the center. For vertical distance larger than $5$ cm, the threshold is $0.4$ cm; for vertical distance between $3$ cm and $5$ cm, the threshold is $0.2$ cm; for vertical distance smaller than $3$ cm, the threshold is $-0.5$ cm (moving away from the center for more than $0.5$ cm in the last step).

    \item The vertical movement of the end-effector towars the surface of the pillow during last step is less than $0.15$ cm for 10 consecutive steps.
\end{enumerate}

During expert control, the expert disengages if the horizontal end-effector towards the center of the pillow during last step is greater than the pre-defined threshold for 3 consecutive steps.

\subsection{Human-in-the-loop Experiments}
\label{app:human_in_the_loop_experiments}

For the human-in-the-loop experiments, we substitute the synthesized experts in the corresponding experiments with human experts.

\subsubsection{Highway-Human}
\label{app:highway_human}

\begin{figure*}[t]
    \centering
    \subcaptionbox{Policy Control
    \label{fig:gui_1}}
    [0.24\linewidth][c]{
        \includegraphics[width=\linewidth]{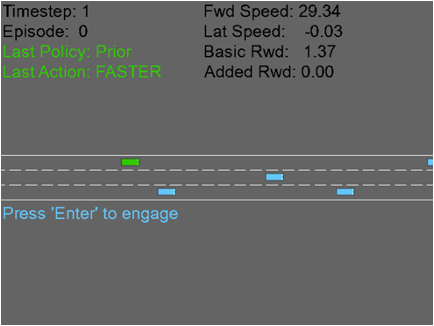}
    }
    \subcaptionbox{Human Engage
    \label{fig:gui_2}}
    [0.24\linewidth][c]{
        \includegraphics[width=\linewidth]{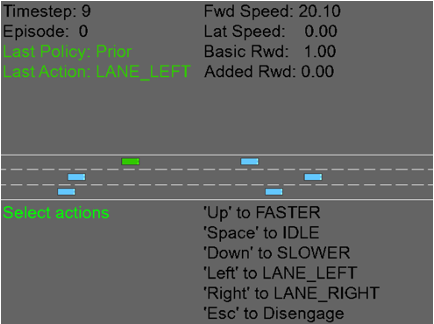}
    }
    \subcaptionbox{Human Control
    \label{fig:gui_3}}
    [0.24\linewidth][c]{
        \includegraphics[width=\linewidth]{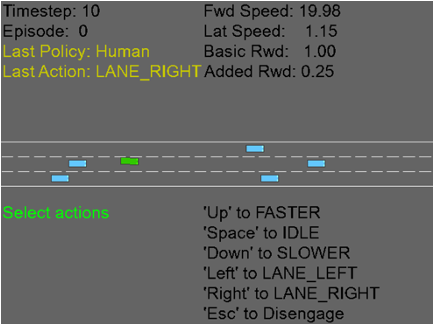}
    }
    \subcaptionbox{Human Disengage
    \label{fig:gui_4}}
    [0.24\linewidth][c]{
        \includegraphics[width=\linewidth]{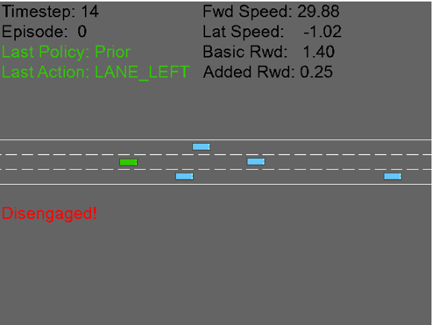}
    }
    \caption{\textbf{\textit{Highway-Human} Graphic User Interface.} There are four different scenarios during the sample collection process. When the human expert engages and takes over the control, additional information would show up for available actions.}
    \label{fig:highway_human_gui}
\end{figure*}

\textbf{Overview.} We use the same \texttt{highway-env} environment with a customized Graphic User Interface (GUI) for human supervision.  Human experts can intervene at will and control the ego vehicle using the keyboard. The sample GUI of 4 different scenarios are shown in Fig.~\ref{fig:highway_human_gui}.

\textbf{Human Interface.}
We design a customized Graphic User Interface (GUI) for \texttt{highway-env} as shown in Fig.~\ref{fig:highway_human_gui}. The upper-left corner contains information about: 1) the step count in the current episode; 2) the total episode count; and 3) last executed action and last policy in control. The upper-right corner contains information about: 1) forward and lateral speed of the ego vehicle; and 2) basic and residual reward of the current state. The lower-left corner contains the user instruction on engaging and action selection. Whenever the human user is taking control, the lower-right corner shows the available actions and the corresponding keys.

\subsubsection{Bottle-Pushing-Human}
\label{app:bottle_pushing_human}

\textbf{Overview.} We use a Fanuc LR Mate 200$i$D/7L 6-DoF robot arm with a customized tooltip to push the bottle. Human experts can intervene at will and control the robot using a 3DConnexion SpaceMouse. Please refer to Fig.~\ref{fig:robot_demo} (left) for a sample failure rollout where the robot knocks down the wine bottle before alignment, and a sample rollout where the robot successfully pushes the bottle to the goal position after alignment. 

\begin{figure*}[t]
    \centering
    \begin{subfigure}[t]{0.48\textwidth}
        \centering
        \includegraphics[height=5cm]{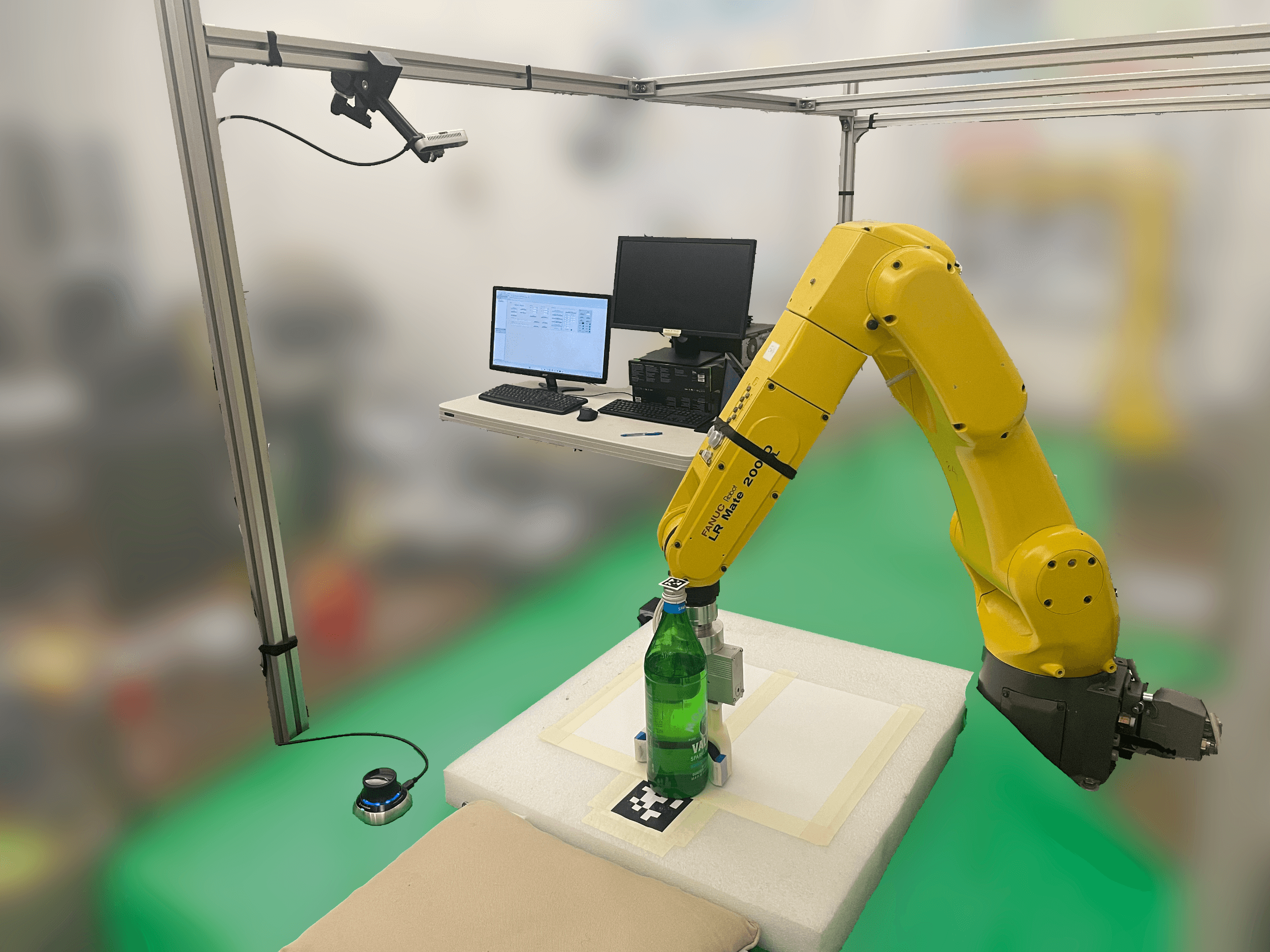}
        \caption{\textit{Bottle-Pushing-Human} Hardware Setup}
        \label{fig:fanuc_hardware_setup}
    \end{subfigure}\hfill
    \begin{subfigure}[t]{0.48\textwidth}
        \centering
        \includegraphics[height=5cm]{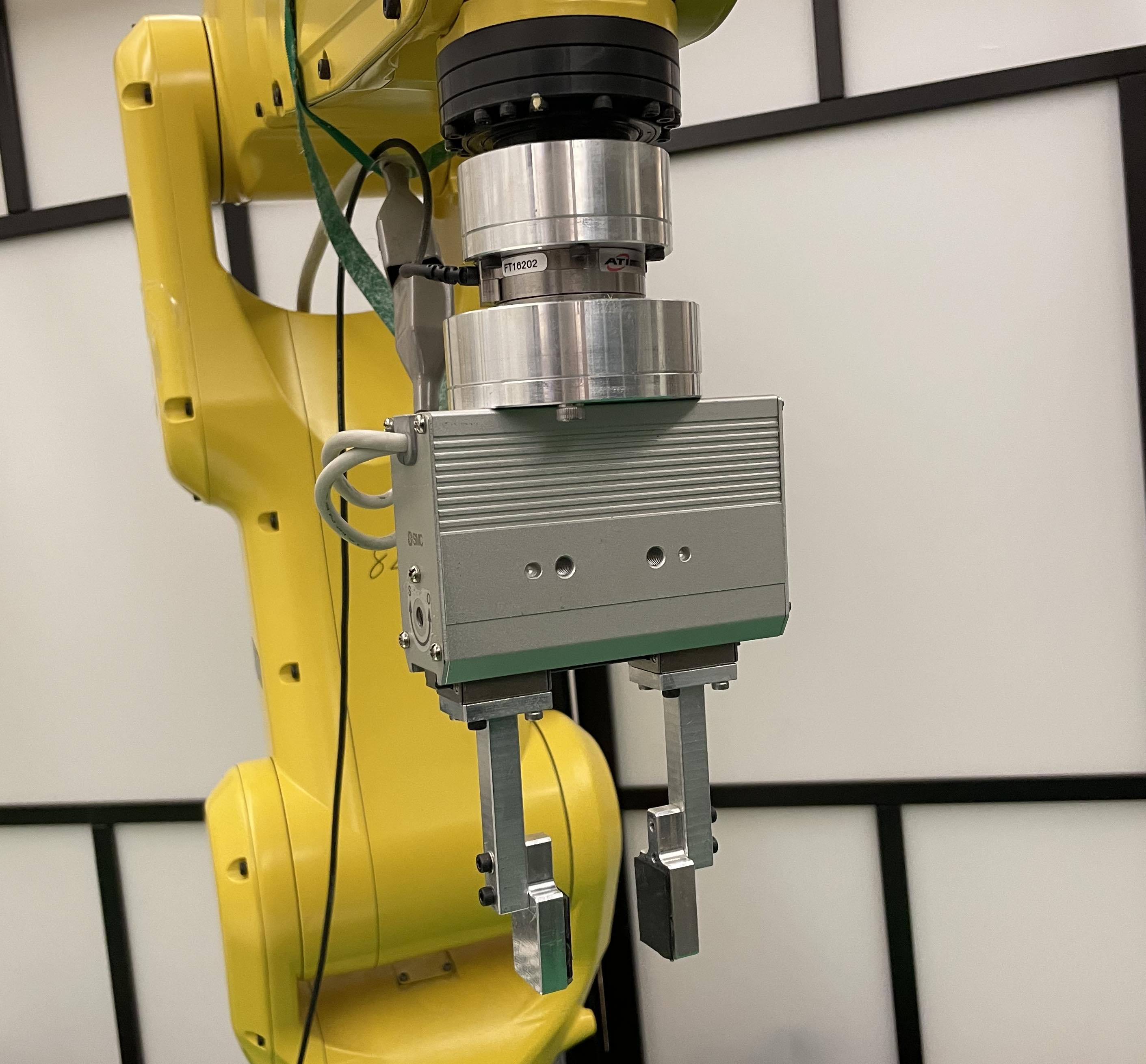}
        \caption{\textit{Pillow-Grasping-Human} Robot Gripper}
        \label{fig:parallel_gripper}
    \end{subfigure}
    \caption{\textbf{Hardware setups for robot experiments.} \textbf{(Left)} In the \textit{Bottle-Pushing-Human} task, we use a Fanuc LR Mate 200$i$D/7L 6-DoF robot arm mounted on a tabletop, a fixed RealSense D435 depth camera for tracking AprilTags attached to the bottle and goal position, and a 3Dconnexion SpaceMouse for online human intervention.  \textbf{(Right)} In the \textit{Pillow-Grasping-Human} task, we use a two-finger parallel gripper mounted on the robot end-effector for grasping the pillow.}
    \label{fig:hardware_setups}
\end{figure*}

\textbf{Human Interface.}
The hardware setup for the real-world experiment is shown in Fig.~\ref{fig:fanuc_hardware_setup}. The robot arm is mounted on the tabletop. We use the RealSense d435 depth camera to track the AprilTags attached to the bottle and the goal position for the state feedback. The human expert uses the SpaceMouse to control the 3D position and orientation of the end-effector. The end-effector consists of a pair of tooltips specifically designed for the bottle-pushing task, which are 3D printed and attached to a parallel gripper with a fixed distance between the two fingers.

\subsubsection{Pillow-Grasping-Human}
\label{app:pillow_grasping_human}

We use the same robot arm with a standard two-finger parallel gripper (see Fig.~\ref{fig:parallel_gripper}) to grasp the pillow. Human experts can intervene at will and control the robot using a 3DConnexion SpaceMouse. Please refer to Fig.~\ref{fig:robot_demo} (right) for a sample failure roll-out where the robot fails to grasp the pillow by the center before alignment, and a sample roll-out where the robot successfully grasps the pillow by the center after alignment. The human interface is the same as \textit{Bottle-Pushing-Human}.

\section{Additional Results}
\label{app:additional_results}
{\bf Sample Efficiency.} Tab.~\ref{tab:sample_efficiency} and Tab.~\ref{tab:human_efforts} present the detailed numerical results corresponding to the plots shown in Fig.~\ref{fig:sample_efficiency} and Fig.~\ref{fig:human_efforts}, respectively. Both tables report the mean values and 95\% confidence intervals of the number of expert samples required by each algorithm. The results clearly demonstrate the advantage of {\textsc{MEReQ}} over the baseline methods with respect to sample efficiency.  

\begin{figure*}[t]
    \centering
    \includegraphics[width=1\linewidth]{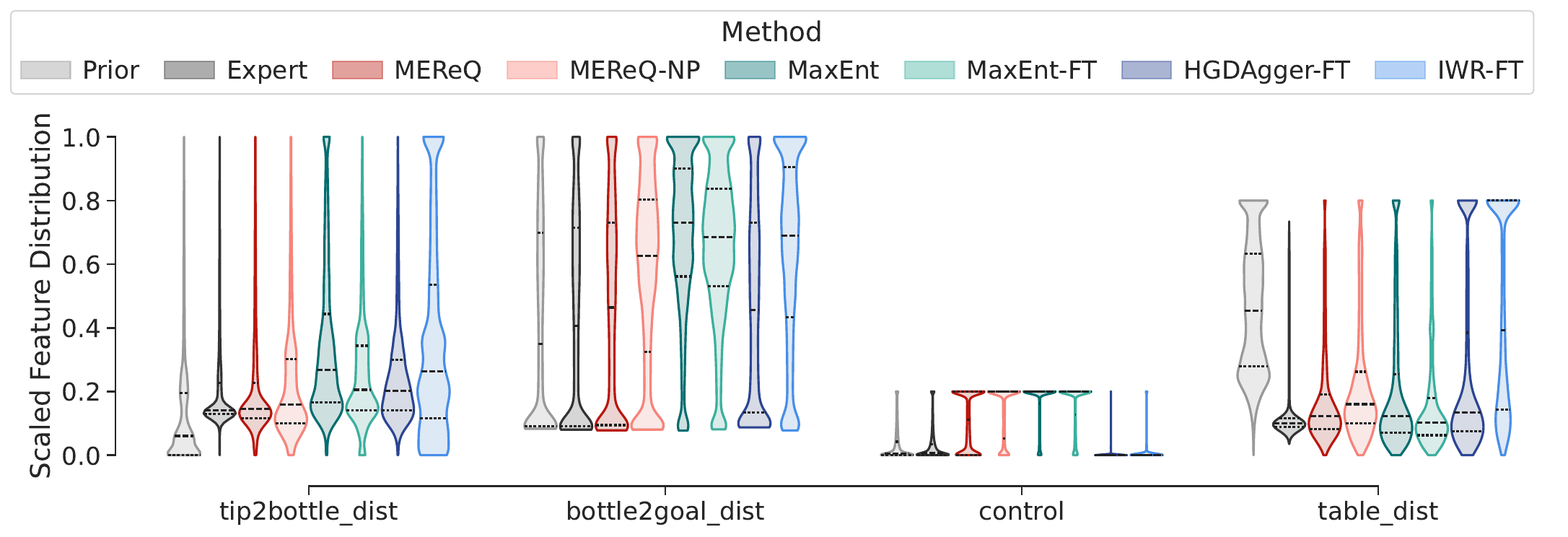}
    \caption{\textbf{Behavior Alignment.} We evaluate the policy distribution of all methods with a convergence threshold of 0.1 for each feature in the \textit{Bottle-Pushing-Sim} environment. All methods align well with the \textcolor{myexpert}{\textbf{Expert}} in the feature \texttt{table\_dist} except for \textcolor{iwr}{\textbf{IWR-FT}}. Additionally, \textcolor{mereq}{\textbf{MEReQ}} aligns better with the \textcolor{myexpert}{\textbf{Expert}} across the other three features compared to other baselines.}
    \label{fig:feature_alignment}
\end{figure*}

{\bf Behavior Alignment.} As discussed in Sec.\ref{sec:simulation_results}, when using a synthesized expert, we can directly \emph{measure the alignment between the behaviors of the learned and expert policies}, since both the expert policy distribution and the ground-truth expert reward are available. Specifically, for the \emph{Bottle-Pushing-Sim} task, we collect sample rollouts from both policies, estimate their feature distributions, and compute the Jensen–Shannon divergence~\citep{menendez1997jensen} between these distributions as a quantitative measure of behavior alignment. The feature distributions and their corresponding Jensen–Shannon divergences relative to the expert policy are shown in Fig.~\ref{fig:feature_alignment} and Tab.~\ref{tab:js_feature}. We also visualize the reward distributions for all policies in Fig.~\ref{fig:feature_alignment} and report their means and standard deviations in Tab.~\ref{tab:reward}. These results show that the \textsc{MEReQ} policy more closely matches the synthesized expert in terms of both feature and reward distributions compared to the baseline methods.

{\bf Performance under Noisy Intervention.} Our work focuses on learning from interventions provided by a single, consistent human expert—\textit{i.e.}, assuming a single trainer whose behavior preference does not shift. There could still be some noise, and indeed that was not controlled for in the experiments. However, handling noisy or inconsistent interventions is beyond our scope and remains a valuable direction for future work. As a preliminary exploration, we introduced Gaussian noise (mean $0$, standard deviation $0.1$) to the normalized actions $[-1,1]$ of synthesized expert interventions (see Tab.~\ref{tab:noisy_expert}, results are reported in \texttt{mean(95\%ci)} with $\delta=0.1$) in the \emph{Bottle-Pushing-Sim} environment. While \textsc{MEReQ}'s performance degrades under injected noise, it still outperforms the baselines. 

\begin{table*}[t]
  \centering
  \caption{\textcolor{mereq}{\textbf{MEReQ}} and its variation \textcolor{mereqnp}{\textbf{MEReQ-NP}} require fewer total expert samples to achieve comparable policy performance compared to the max-ent IRL baselines \textcolor{maxent}{\textbf{MaxEnt}} and \textcolor{maxentft}{\textbf{MaxEnt-FT}}, and interactive imitation learning baselines \textcolor{hgdagger}{\textbf{HG-DAgger-FT}} and \textcolor{iwr}{\textbf{IWR-FT}} under varying criteria strengths in different task and environment. Results are reported in \texttt{mean} (\texttt{95\%ci}).}
  \label{tab:sample_efficiency}
  \begin{adjustbox}{width=\textwidth}
  \begin{tabular}{cccccccc}\toprule
    \textit{Environment} & \textit{$\delta$} & \textcolor{mereq}{\textbf{MEReQ}} & \textcolor{mereqnp}{\textbf{MEReQ-NP}} & \textcolor{maxent}{\textbf{MaxEnt}} & \textcolor{maxentft}{\textbf{MaxEnt-FT}} & \textcolor{hgdagger}{\textbf{HG-DAgger-FT}} & \textcolor{iwr}{\textbf{IWR-FT}}\\
    \midrule
    \multirow{3}{*}{\centering \textbf{Highway-Sim}} & 0.05  &  \textbf{1819 (456)} & 1990 (687) & 4363 (1266) & 4330 (1255) & 1871 (183) & 2284 (1039) \\ 
    & 0.1  &  \textbf{1208 (254)} & 1043 (154) & 2871 (1357) & 1612 (673) & 1754 (160) & 1856 (1214) \\ 
    & 0.15  &  \textbf{965 (100)} & 965 (37) & 2005 (840) & 1336 (468) & 1458 (194) & 1527 (930) \\ 
    \midrule
    \multirow{3}{*}{\centering \textbf{Bottle-Pushing-Sim}} & 0.05  &  \textbf{1707 (261)} & 3338 (1059) & 5298 (2000) & 2976 (933) & 2519 (1459) & 3554 (1118)\\ 
    & 0.1  &  \textbf{1613 (141)} & 2621 (739) & 4536 (1330) & 2636 (468) & 1706 (785) & 2280 (1273)\\ 
    & 0.15  &  1604 (134) & 2159 (717) & 4419 (1306) & 2618 (436) & 1692 (787) & \textbf{1290 (516)}\\ 
    \midrule
    \multirow{3}{*}{\centering \textbf{Erasing-Sim}} & 0.05  &  \textbf{925 (51)} & 989 (228) & 8627 (3019) & 1899 (2796) & 1268 (827) & 4236 (1670)\\ 
    & 0.1  &  \textbf{923 (45)} & 989 (228) & 7965 (3610) & 1899 (2796) & 1258 (842) & 3643 (2231)\\ 
    & 0.15  &  \textbf{923 (45)} & 989 (228) & 7965 (3610) & 1899 (2796) & 1258 (842) & 2968 (1934)\\
    \midrule
    \multirow{3}{*}{\centering \textbf{Pillow-Grasping-Sim}} & 0.05  &  \textbf{2848 (699)} & 3086 (672) & 4992 (2375) & 3188 (1360) & 7699 (624) & 9645 (1034)\\ 
    & 0.1  &  \textbf{2398 (470)} & 2807 (558) & 4127 (2737) & 2808 (1135) & 6490 (1696) & 9645 (1034)\\ 
    & 0.15  &  \textbf{2284 (332)} & 2564 (633) & 3993 (2681) & 2715 (913) & 5427 (2170) & 8879 (1960)\\
    \bottomrule
  \end{tabular}
  \end{adjustbox}
\end{table*}

\begin{table*}[t]
  \centering
  \caption{\textcolor{mereq}{\textbf{MEReQ}} require fewer total human samples to align the prior policy with human preference. Results are reported in \texttt{mean} (\texttt{95\%ci}).}
  \label{tab:human_efforts}
  \begin{adjustbox}{width=\textwidth}
  \begin{tabular}{cccccc}\toprule
    \textit{Environment} & \textcolor{mereq}{\textbf{MEReQ}} & \textcolor{maxent}{\textbf{MaxEnt}} & \textcolor{maxentft}{\textbf{MaxEnt-FT}}
    & \textcolor{hgdagger}{\textbf{HG-DAgger-FT}} & \textcolor{iwr}{\textbf{IWR-FT}}\\
    \midrule
    \textbf{\textit{Highway-Human}} & \textbf{654 (174)} & 2482 (390) & 1270 (440) & 864 (194) & 927 (237) \\
    \textbf{\textit{Bottle-Pushing-Human}}  & \textbf{423 (107)} & 879 (56) & 564 (35) & 450 (105) & 524 (130)\\
    \textbf{\textit{Pillow-Grasping-Human}} & \textbf{149 (20)} & 376 (123) & 234 (141) & 456 (126) & 497 (301)\\ 
    \bottomrule
  \end{tabular}
  \end{adjustbox}
\end{table*}

\begin{table*}[t]
  \centering
  \caption{The Jensen-Shannon Divergence of the feature distribution between each method and the synthesized expert in the \textit{Bottle-Pushing-Sim} environment.  Results are reported in \texttt{mean} (\texttt{95\%ci}). The intervention rate threshold is set to 0.1.}
  \label{tab:feature_alignment}
  \begin{adjustbox}{width=\textwidth}
  \begin{tabular}{cccccccc}\toprule
    \textit{Features} & \textcolor{mereq}{\textbf{MEReQ}} & \textcolor{mereqnp}{\textbf{MEReQ-NP}} & \textcolor{maxent}{\textbf{MaxEnt}} & \textcolor{maxentft}{\textbf{MaxEnt-FT}} & \textcolor{hgdagger}{\textbf{HG-DAgger-FT}} & \textcolor{iwr}{\textbf{IWR-FT}}\\
    \midrule
    \texttt{scaled\_tip2wine}  &  \textbf{0.237 (0.032)} & 0.265 (0.023) & 0.245 (0.022) & 0.250 (0.038) & 0.240 (0.017) & 0.302 (0.058) \\ 
    \texttt{scaled\_wine2goal}  &  \textbf{0.139 (0.005)} & 0.194 (0.044) & 0.247 (0.046) & 0.238 (0.039) & 0.167 (0.033) & 0.236 (0.040) \\ 
    \texttt{scaled\_eef\_acc\_sqrsum}  &  \textbf{0.460 (0.018)} & 0.479 (0.022) & 0.500 (0.026) & 0.505 (0.016) & 0.707 (0.006) & 0.654 (0.022) \\ 
    \texttt{scaled\_table\_dist}  &  \textbf{0.177 (0.021)} & 0.219 (0.025) & 0.236 (0.029) & 0.210 (0.049) & 0.284 (0.080) & 0.308 (0.051) \\ 
    \bottomrule
  \end{tabular}
  \label{tab:js_feature}
  \end{adjustbox}
\end{table*}

\begin{figure}[t]
    \centering
    \includegraphics[width=0.7\linewidth]{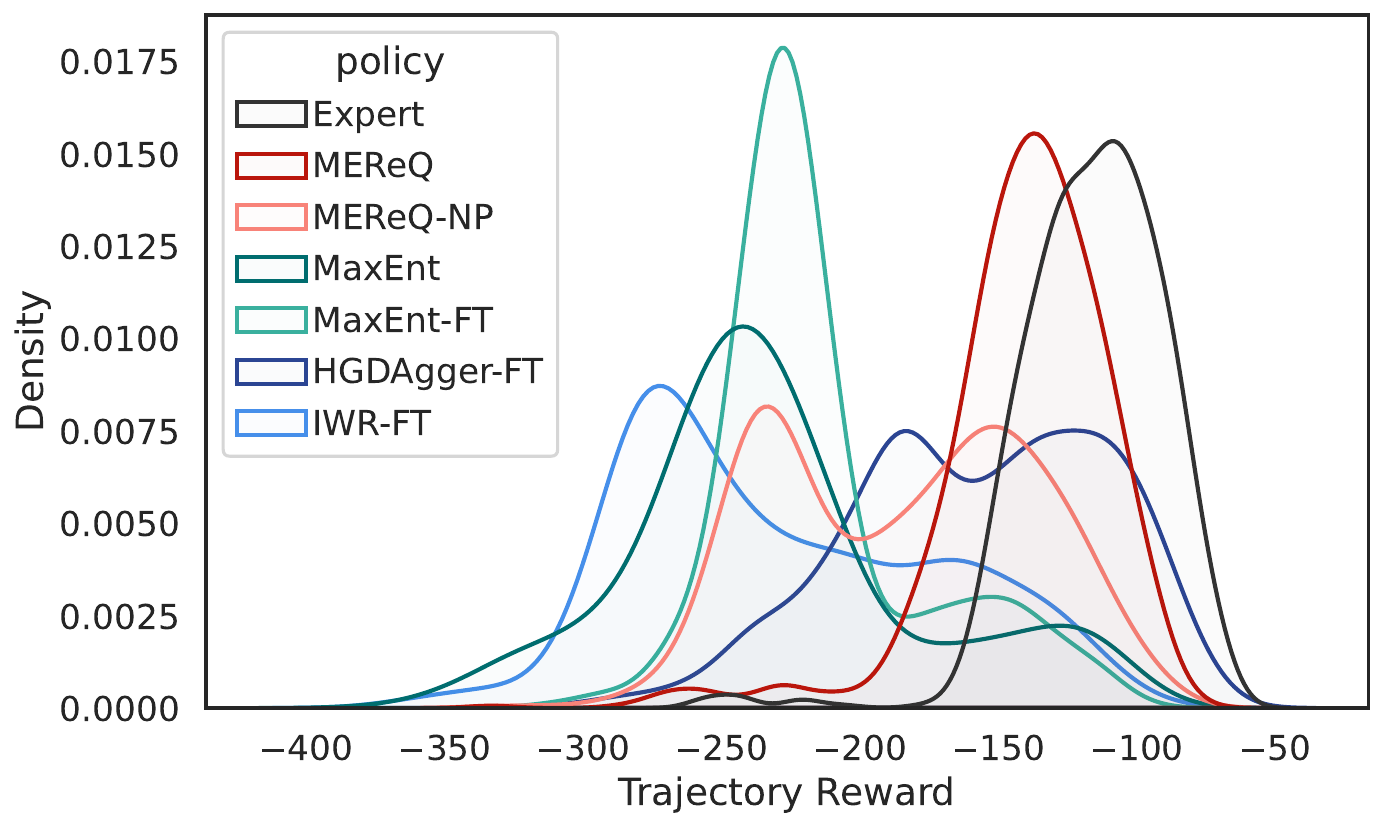}
    \caption{\textbf{Reward Alignment.} We visualize the reward distributions of all methods with a convergence threshold of 0.1 for each feature in the \textit{Bottle-Pushing-Sim} environment. \textcolor{mereq}{\textbf{MEReQ}} aligns best with the \textcolor{myexpert}{\textbf{Expert}} compared to other baselines.}
    \label{fig:rewards}
\end{figure}

\begin{table*}[t]
  \centering
  \caption{The mean and standard deviation of the reward distribution of each method.}
  \label{tab:reward}
  \begin{adjustbox}{width=\textwidth}
  \begin{tabular}{cccccccc}\toprule
    \textcolor{myexpert}{\textbf{Expert}} & \textcolor{mereq}{\textbf{MEReQ}} & \textcolor{mereqnp}{\textbf{MEReQ-NP}} & \textcolor{maxent}{\textbf{MaxEnt}} & \textcolor{maxentft}{\textbf{MaxEnt-FT}} & \textcolor{hgdagger}{\textbf{HG-DAgger-FT}} & \textcolor{iwr}{\textbf{IWR-FT}}\\
    \midrule
    -115.9 (25.9) & \textbf{-140.5 (30.8)} & -184.7 (46.9) & -231.1 (52.9) & -214.1 (36.7) & -157.5 (46.1) & -228.1 (56.1)\\ 
    \bottomrule
  \end{tabular}
  \end{adjustbox}
\end{table*}

\begin{table}[htbp]
    \centering
    \caption{Number of total expert samples with noisy intervention.}
    \begin{tabular}{ccccc}
    \toprule
         & MEReQ & MaxEnt-FT & HG-DAgger-FT & IWR-FT\\
         \midrule
         No Noise & \textbf{1613 (141)} & 2636 (468) & 1706 (785) & 2280 (1273)\\
         10\% Noise & \textbf{1043 (420)} & 2228 (182) & 3987 (1831) & 11921 (1749)\\
         50\% Noise & \textbf{1011 (315)} & 2612 (252) & 3612 (1529) & 11487 (3966)\\
    \bottomrule
    \end{tabular}
    \label{tab:noisy_expert}
\end{table}

\section{Implementation Details}
In this section, we provide the hyperparameters for the prior policy training (see Tab.~\ref{tab:DQN_hyper} and Tab.~\ref{tab:SAC_hyper}) and the Residual Q-Learning training (see Tab.~\ref{tab:Residual_DQN_hyper} and Tab.~\ref{tab:Residual_SAC_hyper}).

\begin{table*}[ht]
    \centering
    \caption{Hyperparameters of DQN Policies.}
    \begin{tabular}{ccc}
    \toprule
        \textit{Hyperparameter} & \textbf{\textit{Highway-Sim}} & \textbf{\textit{Highway-Human}}\\
    \midrule
        \texttt{n\_timesteps} & $5\times 10^5$ & $5\times 10^5$\\
        \texttt{learning\_rate} & $10^{-4}$ & $10^{-4}$\\
        \texttt{batch\_size} & $32$ & $32$\\
        \texttt{buffer\_size} & $1.5\times 10^4$ & $1.5\times 10^4$\\
        \texttt{learning\_starts} & $200$ & $200$\\
        \texttt{gamma} & $0.8$ & $0.8$\\
        \texttt{target\_update\_interval} & $50$ & $50$\\
        \texttt{train\_freq} & $1$ & $1$\\
        \texttt{gradient\_steps} & $1$ & $1$\\
        \texttt{exploration\_fraction} & $0.7$ & $0.7$\\
        \texttt{net\_arch} & $[256, 256]$ & $[256, 256]$\\
    \bottomrule
    \end{tabular}
    \label{tab:DQN_hyper}
\end{table*}

\begin{table*}[ht]
    \centering
    \caption{Hyperparameters of Residual DQN Policies.}
    \begin{tabular}{ccc}
    \toprule
        \textit{Hyperparameter} & \textbf{\textit{Highway-Sim}} & \textbf{\textit{Highway-Human}}\\
    \midrule
        \texttt{n\_timesteps} & $4\times 10^4$ & $4\times 10^4$\\
        \texttt{batch\_size} & $32$ & $32$\\
        \texttt{buffer\_size} & $2000$ & $2000$\\
        \texttt{learning\_starts} & $2000$ & $2000$\\
        \texttt{learning\_rate} & $10^{-4}$ & $10^{-4}$\\
        \texttt{gamma} & $0.8$ & $0.8$\\
        \texttt{target\_update\_interval} & $50$ & $50$\\
        \texttt{train\_freq} & $1$ & $1$\\
        \texttt{gradient\_steps} & $1$ & $1$\\
        \texttt{exploration\_fraction} & $0.7$ & $0.7$\\
        \texttt{net\_arch} & $[256, 256]$ & $[256, 256]$\\
        \texttt{env\_update\_freq} & $1000$ & $1000$\\
        \texttt{sample\_length} & $1000$ & $1000$\\
        \texttt{epsilon} & $0.03$ & $0.03$\\
        \texttt{eta} & $0.2$ & $0.2$\\
    \bottomrule
    \end{tabular}
    \label{tab:Residual_DQN_hyper}
\end{table*}

\begin{table*}[ht]
    \centering
    \caption{Hyperparameters of SAC Policies.}
    \begin{adjustbox}{width=\textwidth}    
    \begin{tabular}{cccccc}
    \toprule
        \textit{Hyperparameter} & \textbf{\textit{Bottle-Pushing-Sim}} & \textbf{\textit{Bottle-Pushing-Human}} & \textbf{\textit{Erasing-Sim}} & \textbf{\textit{Pillow-Grasping-Sim}} & \textbf{\textit{Pillow-Grasping-Human}}\\
    \midrule
        \texttt{n\_timesteps} & $5\times 10^4$ & $5\times 10^4$ & $5\times 10^4$ & $5\times 10^4$ & $5\times 10^4$\\
        \texttt{learning\_rate} & $5\times10^{-3}$ & $5\times10^{-3}$ & $5\times10^{-3}$ & $5\times10^{-3}$ & $5\times10^{-3}$\\
        \texttt{batch\_size} & $512$ & $512$ & $512$ & $512$ & $512$\\
        \texttt{buffer\_size} & $10^6$ & $10^6$ & $10^6$ & $10^6$ & $10^6$\\
        \texttt{learning\_starts} & $5000$ & $5000$ & $5000$ & $5000$ & $5000$\\
        \texttt{ent\_coef} & \texttt{auto} & \texttt{auto} & \texttt{auto} & \texttt{auto} & \texttt{auto}\\
        \texttt{gamma} & $0.9$ & $0.9$ & $0.9$ & $0.9$ & $0.9$\\
        \texttt{tau} & $0.01$ & $0.01$ & $0.01$ & $0.01$ & $0.01$\\
        \texttt{train\_freq} & $1$ & $1$ & $1$ & $1$ & $1$\\
        \texttt{gradient\_steps} & $1$ & $1$ & $1$ & $1$ & $1$\\
        \texttt{net\_arch} & $[400, 300]$ & $[400, 300]$ & $[400, 300]$ & $[400, 300]$ & $[400, 300]$\\
    \bottomrule
    \end{tabular}
    \end{adjustbox}
    \label{tab:SAC_hyper}
\end{table*}

\begin{table*}[ht]
    \centering
    \caption{Hyperparameters of Residual SAC Policies.}
    \begin{adjustbox}{width=\textwidth}
    \begin{tabular}{cccccc}
    \toprule
        \textit{Hyperparameter} & \textbf{\textit{Bottle-Pushing-Sim}} & \textbf{\textit{Bottle-Pushing-Human}} & \textbf{\textit{Erasing-Sim}} & \textbf{\textit{Pillow-Grasping-Sim}} & \textbf{\textit{Pillow-Grasping-Human}}\\
    \midrule
        \texttt{n\_timesteps} & $2\times 10^4$ & $2\times 10^4$ & $2\times 10^4$ & $2\times 10^4$ & $2\times 10^4$\\
        \texttt{batch\_size} & $512$ & $512$ & $512$ & $512$ & $512$\\
        \texttt{buffer\_size} & $10^6$ & $10^6$ & $10^6$ & $10^6$ & $10^6$\\
        \texttt{learning\_starts} & $5000$ & $5000$ & $5000$ & $5000$ & $5000$\\
        \texttt{learning\_rate} & $5\times10^{-3}$ & $5\times10^{-3}$ & $5\times10^{-3}$ & $5\times10^{-3}$ & $5\times10^{-3}$\\
        \texttt{ent\_coef} & \texttt{auto} & \texttt{auto} & \texttt{auto} & \texttt{auto} & \texttt{auto}\\
        \texttt{ent\_coef\_prior} & $0.035$ & $0.035$ & $0.035$ & $0.035$ & $0.035$\\
        \texttt{gamma} & $0.9$ & $0.9$ & $0.9$ & $0.9$ & $0.9$\\
        \texttt{tau} & $0.01$ & $0.01$ & $0.01$ & $0.01$ & $0.01$\\
        \texttt{train\_freq} & $1$ & $1$ & $1$ & $1$ & $1$\\
        \texttt{gradient\_steps} & $1$ & $1$ & $1$ & $1$ & $1$\\
        \texttt{net\_arch} & $[400, 300]$ & $[400, 300]$ & $[400, 300]$ & $[400, 300]$ & $[400, 300]$\\
        \texttt{env\_update\_freq} & $1000$ & $1000$ & $1000$ & $1000$ & $1000$\\
        \texttt{sample\_length} & $1000$ & $1000$ & $2000$ & $2000$ & $1000$\\
        \texttt{epsilon} & $0.2$ & $0.2$ & $0.1$ & $0.1$ & $0.4$\\
        \texttt{eta} & $0.2$ & $0.2$ & $0.2$ & $0.2$ & $0.2$\\
    \bottomrule
    \end{tabular}
    \end{adjustbox}
    \label{tab:Residual_SAC_hyper}
\end{table*}

%% file: root.bbl
\begin{thebibliography}{56}
\providecommand{\natexlab}[1]{#1}
\providecommand{\url}[1]{\texttt{#1}}
\expandafter\ifx\csname urlstyle\endcsname\relax
  \providecommand{\doi}[1]{doi: #1}\else
  \providecommand{\doi}{doi: \begingroup \urlstyle{rm}\Url}\fi

\bibitem[Ji et~al.(2023)Ji, Qiu, Chen, Zhang, Lou, Wang, Duan, He, Zhou, Zhang, et~al.]{ji2023ai}
J.~Ji, T.~Qiu, B.~Chen, B.~Zhang, H.~Lou, K.~Wang, Y.~Duan, Z.~He, J.~Zhou, Z.~Zhang, et~al.
\newblock Ai alignment: A comprehensive survey.
\newblock \emph{arXiv preprint arXiv:2310.19852}, 2023.

\bibitem[Arzate~Cruz and Igarashi(2020)]{arzate2020survey}
C.~Arzate~Cruz and T.~Igarashi.
\newblock A survey on interactive reinforcement learning: Design principles and open challenges.
\newblock In \emph{Proceedings of the 2020 ACM designing interactive systems conference}, pages 1195--1209, 2020.

\bibitem[Cui et~al.(2021)Cui, Koppol, Admoni, Niekum, Simmons, Steinfeld, and Fitzgerald]{cui2021understanding}
Y.~Cui, P.~Koppol, H.~Admoni, S.~Niekum, R.~Simmons, A.~Steinfeld, and T.~Fitzgerald.
\newblock Understanding the relationship between interactions and outcomes in human-in-the-loop machine learning.
\newblock In \emph{International Joint Conference on Artificial Intelligence}, 2021.

\bibitem[Kelly et~al.(2019)Kelly, Sidrane, Driggs-Campbell, and Kochenderfer]{kelly2019hg}
M.~Kelly, C.~Sidrane, K.~Driggs-Campbell, and M.~J. Kochenderfer.
\newblock Hg-dagger: Interactive imitation learning with human experts.
\newblock In \emph{2019 International Conference on Robotics and Automation (ICRA)}, pages 8077--8083. IEEE, 2019.

\bibitem[Liu et~al.(2023)Liu, Nasiriany, Zhang, Bao, and Zhu]{liu2023robot}
H.~Liu, S.~Nasiriany, L.~Zhang, Z.~Bao, and Y.~Zhu.
\newblock Robot learning on the job: Human-in-the-loop autonomy and learning during deployment.
\newblock \emph{Robotics: Science and Systems (R:SS)}, 2023.

\bibitem[Zhang and Cho(2016)]{zhang2016query}
J.~Zhang and K.~Cho.
\newblock Query-efficient imitation learning for end-to-end autonomous driving.
\newblock \emph{arXiv e-prints}, pages arXiv--1605, 2016.

\bibitem[Ross and Bagnell(2010)]{ross2010efficient}
S.~Ross and D.~Bagnell.
\newblock Efficient reductions for imitation learning.
\newblock In \emph{Proceedings of the thirteenth international conference on artificial intelligence and statistics}, pages 661--668. JMLR Workshop and Conference Proceedings, 2010.

\bibitem[Garg et~al.(2021)Garg, Chakraborty, Cundy, Song, and Ermon]{garg2021iq}
D.~Garg, S.~Chakraborty, C.~Cundy, J.~Song, and S.~Ermon.
\newblock Iq-learn: Inverse soft-q learning for imitation.
\newblock \emph{Advances in Neural Information Processing Systems}, 34:\penalty0 4028--4039, 2021.

\bibitem[Jiang et~al.(2024)Jiang, Wang, Zhang, Wu, and Fei-Fei]{jiang2024transic}
Y.~Jiang, C.~Wang, R.~Zhang, J.~Wu, and L.~Fei-Fei.
\newblock Transic: Sim-to-real policy transfer by learning from online correction.
\newblock \emph{arXiv preprint arXiv:2405.10315}, 2024.

\bibitem[NG(2000)]{ng2000algorithms}
A.~NG.
\newblock Algorithms for inverse reinforcement learning.
\newblock In \emph{Proc. of 17th International Conference on Machine Learning, 2000}, pages 663--670, 2000.

\bibitem[Ziebart et~al.(2008)Ziebart, Maas, Bagnell, and Dey]{ziebart2008maximum}
B.~D. Ziebart, A.~Maas, J.~A. Bagnell, and A.~K. Dey.
\newblock Maximum entropy inverse reinforcement learning.
\newblock In \emph{Proceedings of the 23rd national conference on Artificial intelligence-Volume 3}, pages 1433--1438, 2008.

\bibitem[Arora and Doshi(2021)]{arora2021survey}
S.~Arora and P.~Doshi.
\newblock A survey of inverse reinforcement learning: Challenges, methods and progress.
\newblock \emph{Artificial Intelligence}, 297:\penalty0 103500, 2021.

\bibitem[Von~Neumann and Morgenstern(1947)]{von1947theory}
J.~Von~Neumann and O.~Morgenstern.
\newblock \emph{Theory of games and economic behavior, 2nd rev}.
\newblock Princeton university press, 1947.

\bibitem[Baker et~al.(2007)Baker, Tenenbaum, and Saxe]{baker2007goal}
C.~L. Baker, J.~B. Tenenbaum, and R.~R. Saxe.
\newblock Goal inference as inverse planning.
\newblock In \emph{Proceedings of the annual meeting of the cognitive science society}, volume~29, 2007.

\bibitem[Li et~al.(2024)Li, Tang, Nishimura, Mercat, Tomizuka, and Zhan]{li2024residual}
C.~Li, C.~Tang, H.~Nishimura, J.~Mercat, M.~Tomizuka, and W.~Zhan.
\newblock Residual q-learning: Offline and online policy customization without value.
\newblock \emph{Advances in Neural Information Processing Systems}, 36, 2024.

\bibitem[Yue et~al.(2012)Yue, Broder, Kleinberg, and Joachims]{yue2012k}
Y.~Yue, J.~Broder, R.~Kleinberg, and T.~Joachims.
\newblock The k-armed dueling bandits problem.
\newblock \emph{Journal of Computer and System Sciences}, 78\penalty0 (5):\penalty0 1538--1556, 2012.

\bibitem[Jain et~al.(2013)Jain, Wojcik, Joachims, and Saxena]{jain2013learning}
A.~Jain, B.~Wojcik, T.~Joachims, and A.~Saxena.
\newblock Learning trajectory preferences for manipulators via iterative improvement.
\newblock \emph{Advances in neural information processing systems}, 26, 2013.

\bibitem[Christiano et~al.(2017)Christiano, Leike, Brown, Martic, Legg, and Amodei]{christiano2017deep}
P.~F. Christiano, J.~Leike, T.~Brown, M.~Martic, S.~Legg, and D.~Amodei.
\newblock Deep reinforcement learning from human preferences.
\newblock \emph{Advances in neural information processing systems}, 30, 2017.

\bibitem[B{\i}y{\i}k et~al.(2022)B{\i}y{\i}k, Losey, Palan, Landolfi, Shevchuk, and Sadigh]{biyik2022learning}
E.~B{\i}y{\i}k, D.~P. Losey, M.~Palan, N.~C. Landolfi, G.~Shevchuk, and D.~Sadigh.
\newblock Learning reward functions from diverse sources of human feedback: Optimally integrating demonstrations and preferences.
\newblock \emph{The International Journal of Robotics Research}, 41\penalty0 (1):\penalty0 45--67, 2022.

\bibitem[Lee et~al.(2021)Lee, Smith, and Abbeel]{lee2021pebble}
K.~Lee, L.~Smith, and P.~Abbeel.
\newblock Pebble: Feedback-efficient interactive reinforcement learning via relabeling experience and unsupervised pre-training.
\newblock In \emph{38th International Conference on Machine Learning, ICML 2021}. International Machine Learning Society (IMLS), 2021.

\bibitem[Wang et~al.(2022)Wang, Lee, Hakhamaneshi, Abbeel, and Laskin]{wang2022skill}
X.~Wang, K.~Lee, K.~Hakhamaneshi, P.~Abbeel, and M.~Laskin.
\newblock Skill preferences: Learning to extract and execute robotic skills from human feedback.
\newblock In \emph{Conference on Robot Learning}, pages 1259--1268. PMLR, 2022.

\bibitem[Ouyang et~al.(2022)Ouyang, Wu, Jiang, Almeida, Wainwright, Mishkin, Zhang, Agarwal, Slama, Ray, et~al.]{ouyang2022training}
L.~Ouyang, J.~Wu, X.~Jiang, D.~Almeida, C.~Wainwright, P.~Mishkin, C.~Zhang, S.~Agarwal, K.~Slama, A.~Ray, et~al.
\newblock Training language models to follow instructions with human feedback.
\newblock \emph{Advances in neural information processing systems}, 35:\penalty0 27730--27744, 2022.

\bibitem[Myers et~al.(2023)Myers, B{\i}y{\i}k, and Sadigh]{myers2023active}
V.~Myers, E.~B{\i}y{\i}k, and D.~Sadigh.
\newblock Active reward learning from online preferences.
\newblock In \emph{2023 IEEE International Conference on Robotics and Automation (ICRA)}, pages 7511--7518. IEEE, 2023.

\bibitem[Rafailov et~al.(2024)Rafailov, Sharma, Mitchell, Manning, Ermon, and Finn]{rafailov2024direct}
R.~Rafailov, A.~Sharma, E.~Mitchell, C.~D. Manning, S.~Ermon, and C.~Finn.
\newblock Direct preference optimization: Your language model is secretly a reward model.
\newblock \emph{Advances in Neural Information Processing Systems}, 36, 2024.

\bibitem[Hejna et~al.(2023)Hejna, Rafailov, Sikchi, Finn, Niekum, Knox, and Sadigh]{hejna2023contrastive}
J.~Hejna, R.~Rafailov, H.~Sikchi, C.~Finn, S.~Niekum, W.~B. Knox, and D.~Sadigh.
\newblock Contrastive preference learning: Learning from human feedback without reinforcement learning.
\newblock In \emph{The Twelfth International Conference on Learning Representations}, 2023.

\bibitem[Tian et~al.(2023)Tian, Xu, Tomizuka, Malik, and Bajcsy]{tian2023matters}
T.~Tian, C.~Xu, M.~Tomizuka, J.~Malik, and A.~Bajcsy.
\newblock What matters to you? towards visual representation alignment for robot learning.
\newblock In \emph{The Twelfth International Conference on Learning Representations}, 2023.

\bibitem[Saunders et~al.(2018)Saunders, Sastry, Stuhlm{\"u}ller, and Evans]{saunders2017trial}
W.~Saunders, G.~Sastry, A.~Stuhlm{\"u}ller, and O.~Evans.
\newblock Trial without error: Towards safe reinforcement learning via human intervention.
\newblock In \emph{Proceedings of the 17th International Conference on Autonomous Agents and MultiAgent Systems}, pages 2067--2069, 2018.

\bibitem[Wang et~al.(2021)Wang, Xiao, Liu, Warnell, and Stone]{wang2021appli}
Z.~Wang, X.~Xiao, B.~Liu, G.~Warnell, and P.~Stone.
\newblock Appli: Adaptive planner parameter learning from interventions.
\newblock In \emph{2021 IEEE international conference on robotics and automation (ICRA)}, pages 6079--6085. IEEE, 2021.

\bibitem[Celemin and Ruiz-del Solar(2019)]{celemin2019interactive}
C.~Celemin and J.~Ruiz-del Solar.
\newblock An interactive framework for learning continuous actions policies based on corrective feedback.
\newblock \emph{Journal of Intelligent \& Robotic Systems}, 95:\penalty0 77--97, 2019.

\bibitem[Peng et~al.(2024)Peng, Mo, Duan, Li, and Zhou]{peng2024learning}
Z.~M. Peng, W.~Mo, C.~Duan, Q.~Li, and B.~Zhou.
\newblock Learning from active human involvement through proxy value propagation.
\newblock \emph{Advances in neural information processing systems}, 36, 2024.

\bibitem[Mandlekar et~al.(2020)Mandlekar, Xu, Mart{\'\i}n-Mart{\'\i}n, Zhu, Fei-Fei, and Savarese]{mandlekar2020human}
A.~Mandlekar, D.~Xu, R.~Mart{\'\i}n-Mart{\'\i}n, Y.~Zhu, L.~Fei-Fei, and S.~Savarese.
\newblock Human-in-the-loop imitation learning using remote teleoperation.
\newblock \emph{arXiv preprint arXiv:2012.06733}, 2020.

\bibitem[Spencer et~al.(2020)Spencer, Choudhury, Barnes, Schmittle, Chiang, Ramadge, and Srinivasa]{spencer2020learning}
J.~Spencer, S.~Choudhury, M.~Barnes, M.~Schmittle, M.~Chiang, P.~Ramadge, and S.~Srinivasa.
\newblock Learning from interventions: Human-robot interaction as both explicit and implicit feedback.
\newblock In \emph{16th Robotics: Science and Systems, RSS 2020}. MIT Press Journals, 2020.

\bibitem[Knox and Stone(2012)]{knox2012reinforcement}
W.~B. Knox and P.~Stone.
\newblock Reinforcement learning from human reward: Discounting in episodic tasks.
\newblock In \emph{2012 IEEE RO-MAN: The 21st IEEE international symposium on robot and human interactive communication}, pages 878--885. IEEE, 2012.

\bibitem[Argall et~al.(2010)Argall, Sauser, and Billard]{argall2010tactile}
B.~D. Argall, E.~L. Sauser, and A.~G. Billard.
\newblock Tactile guidance for policy refinement and reuse.
\newblock In \emph{2010 IEEE 9th International Conference on Development and Learning}, pages 7--12. IEEE, 2010.

\bibitem[Fitzgerald et~al.(2019)Fitzgerald, Short, Goel, and Thomaz]{fitzgerald2019human}
T.~Fitzgerald, E.~Short, A.~Goel, and A.~Thomaz.
\newblock Human-guided trajectory adaptation for tool transfer.
\newblock In \emph{Proceedings of the 18th International Conference on Autonomous Agents and MultiAgent Systems}, pages 1350--1358, 2019.

\bibitem[Bajcsy et~al.(2017)Bajcsy, Losey, O’malley, and Dragan]{bajcsy2017learning}
A.~Bajcsy, D.~P. Losey, M.~K. O’malley, and A.~D. Dragan.
\newblock Learning robot objectives from physical human interaction.
\newblock In \emph{Conference on robot learning}, pages 217--226. PMLR, 2017.

\bibitem[Najar et~al.(2020)Najar, Sigaud, and Chetouani]{najar2020interactively}
A.~Najar, O.~Sigaud, and M.~Chetouani.
\newblock Interactively shaping robot behaviour with unlabeled human instructions.
\newblock \emph{Autonomous Agents and Multi-Agent Systems}, 34\penalty0 (2):\penalty0 35, 2020.

\bibitem[Wilde et~al.(2021)Wilde, Biyik, Sadigh, and Smith]{wilde2021learning}
N.~Wilde, E.~Biyik, D.~Sadigh, and S.~L. Smith.
\newblock Learning reward functions from scale feedback.
\newblock In \emph{5th Annual Conference on Robot Learning}, 2021.

\bibitem[Warnell et~al.(2018)Warnell, Waytowich, Lawhern, and Stone]{warnell2018deep}
G.~Warnell, N.~Waytowich, V.~Lawhern, and P.~Stone.
\newblock Deep tamer: Interactive agent shaping in high-dimensional state spaces.
\newblock In \emph{Proceedings of the AAAI conference on artificial intelligence}, volume~32, 2018.

\bibitem[MacGlashan et~al.(2017)MacGlashan, Ho, Loftin, Peng, Wang, Roberts, Taylor, and Littman]{macglashan2017interactive}
J.~MacGlashan, M.~K. Ho, R.~Loftin, B.~Peng, G.~Wang, D.~L. Roberts, M.~E. Taylor, and M.~L. Littman.
\newblock Interactive learning from policy-dependent human feedback.
\newblock In \emph{International conference on machine learning}, pages 2285--2294. PMLR, 2017.

\bibitem[Brown et~al.(2019)Brown, Goo, Nagarajan, and Niekum]{brown2019extrapolating}
D.~Brown, W.~Goo, P.~Nagarajan, and S.~Niekum.
\newblock Extrapolating beyond suboptimal demonstrations via inverse reinforcement learning from observations.
\newblock In \emph{International conference on machine learning}, pages 783--792. PMLR, 2019.

\bibitem[Cui and Niekum(2018)]{cui2018active}
Y.~Cui and S.~Niekum.
\newblock Active reward learning from critiques.
\newblock In \emph{2018 IEEE international conference on robotics and automation (ICRA)}, pages 6907--6914. IEEE, 2018.

\bibitem[Spencer et~al.(2022)Spencer, Choudhury, Barnes, Schmittle, Chiang, Ramadge, and Srinivasa]{spencer2022expert}
J.~Spencer, S.~Choudhury, M.~Barnes, M.~Schmittle, M.~Chiang, P.~Ramadge, and S.~Srinivasa.
\newblock Expert intervention learning: An online framework for robot learning from explicit and implicit human feedback.
\newblock \emph{Autonomous Robots}, pages 1--15, 2022.

\bibitem[Bobu et~al.(2018)Bobu, Bajcsy, Fisac, and Dragan]{bobu2018learning}
A.~Bobu, A.~Bajcsy, J.~F. Fisac, and A.~D. Dragan.
\newblock Learning under misspecified objective spaces.
\newblock In \emph{Conference on Robot Learning}, pages 796--805. PMLR, 2018.

\bibitem[Bobu et~al.(2021)Bobu, Wiggert, Tomlin, and Dragan]{bobu2021feature}
A.~Bobu, M.~Wiggert, C.~Tomlin, and A.~D. Dragan.
\newblock Feature expansive reward learning: Rethinking human input.
\newblock In \emph{Proceedings of the 2021 ACM/IEEE International Conference on Human-Robot Interaction}, pages 216--224, 2021.

\bibitem[Haarnoja et~al.(2017)Haarnoja, Tang, Abbeel, and Levine]{haarnoja2017reinforcement}
T.~Haarnoja, H.~Tang, P.~Abbeel, and S.~Levine.
\newblock Reinforcement learning with deep energy-based policies.
\newblock In \emph{International conference on machine learning}, pages 1352--1361. PMLR, 2017.

\bibitem[Haarnoja et~al.(2018)Haarnoja, Zhou, Abbeel, and Levine]{haarnoja2018soft}
T.~Haarnoja, A.~Zhou, P.~Abbeel, and S.~Levine.
\newblock Soft actor-critic: Off-policy maximum entropy deep reinforcement learning with a stochastic actor.
\newblock In \emph{International conference on machine learning}, pages 1861--1870. PMLR, 2018.

\bibitem[Jaynes(1957)]{jaynes1957information}
E.~T. Jaynes.
\newblock Information theory and statistical mechanics.
\newblock \emph{Physical review}, 106\penalty0 (4):\penalty0 620, 1957.

\bibitem[Bobu et~al.(2020)Bobu, Scobee, Fisac, Sastry, and Dragan]{bobu2020less}
A.~Bobu, D.~R. Scobee, J.~F. Fisac, S.~S. Sastry, and A.~D. Dragan.
\newblock Less is more: Rethinking probabilistic models of human behavior.
\newblock In \emph{Proceedings of the 2020 acm/ieee international conference on human-robot interaction}, pages 429--437, 2020.

\bibitem[Leurent(2018)]{highway-env}
E.~Leurent.
\newblock An environment for autonomous driving decision-making.
\newblock \url{https://github.com/eleurent/highway-env}, 2018.

\bibitem[Todorov et~al.(2012)Todorov, Erez, and Tassa]{todorov2012mujoco}
E.~Todorov, T.~Erez, and Y.~Tassa.
\newblock Mujoco: A physics engine for model-based control.
\newblock In \emph{2012 IEEE/RSJ international conference on intelligent robots and systems}, pages 5026--5033. IEEE, 2012.

\bibitem[Levine(2011)]{levine2011nonlinear}
S.~Levine.
\newblock Nonlinear inverse reinforcement learning with gaussian processes.
\newblock \emph{NeurIPS}, 2011.

\bibitem[Boularias(2011)]{boularias2011relative}
A.~Boularias.
\newblock Relative entropy inverse reinforcement learning.
\newblock In \emph{PMLR}, 2011.

\bibitem[Mnih et~al.(2013)Mnih, Kavukcuoglu, Silver, Graves, Antonoglou, Wierstra, and Riedmiller]{mnih2013playing}
V.~Mnih, K.~Kavukcuoglu, D.~Silver, A.~Graves, I.~Antonoglou, D.~Wierstra, and M.~Riedmiller.
\newblock Playing atari with deep reinforcement learning.
\newblock \emph{arXiv preprint arXiv:1312.5602}, 2013.

\bibitem[Brockman et~al.(2016)Brockman, Cheung, Pettersson, Schneider, Schulman, Tang, and Zaremba]{brockman2016openai}
G.~Brockman, V.~Cheung, L.~Pettersson, J.~Schneider, J.~Schulman, J.~Tang, and W.~Zaremba.
\newblock Openai gym.
\newblock \emph{arXiv preprint arXiv:1606.01540}, 2016.

\bibitem[Men{\'e}ndez et~al.(1997)Men{\'e}ndez, Pardo, Pardo, and Pardo]{menendez1997jensen}
M.~Men{\'e}ndez, J.~Pardo, L.~Pardo, and M.~Pardo.
\newblock The jensen-shannon divergence.
\newblock \emph{Journal of the Franklin Institute}, 334\penalty0 (2):\penalty0 307--318, 1997.

\end{thebibliography}
